\documentclass[10pt]{article}

\usepackage{fullpage}
\usepackage{setspace}
\usepackage{parskip}
\usepackage{titlesec}
\usepackage[section]{placeins}
\usepackage{breakcites}
\usepackage{lineno}
\usepackage{hyphenat}

\PassOptionsToPackage{hyphens}{url}
\PassOptionsToPackage{dvipsnames,table,xcdraw}{xcolor}
\usepackage[colorlinks = true,
            linkcolor = blue,
            urlcolor  = blue,
            citecolor = blue,
            anchorcolor = blue]{hyperref}
\usepackage{etoolbox}

\renewenvironment{abstract}
{\section*{Abstract}}
{\par\bigskip}

\titlespacing{\section}{0pt}{*4}{*1.5}
\titlespacing{\subsection}{0pt}{*2.5}{*1}
\titlespacing{\subsubsection}{0pt}{*1.8}{*0.7}

\usepackage{authblk}

\usepackage{graphicx}
\usepackage[space]{grffile}
\usepackage{latexsym}
\usepackage{textcomp}
\usepackage{longtable}
\usepackage{tabulary}
\usepackage{booktabs,array,multirow}
\usepackage{amsfonts,amsmath,amssymb}
\providecommand\citet{\cite}
\providecommand\citep{\cite}

\newif\iflatexml\latexmlfalse

\AtBeginDocument{\DeclareGraphicsExtensions{.pdf,.PDF,.eps,.EPS,.png,.PNG,.tif,.TIF,.jpg,.JPG,.jpeg,.JPEG}}

\usepackage[utf8]{inputenc}
\usepackage[english]{babel}

\usepackage{float}

\usepackage{subcaption}
\usepackage{tikz}
\usetikzlibrary{shadows,positioning, trees, decorations, calc}
\usepackage[font=small]{caption}
\usepackage{pifont}
\usepackage{makecell}
\usepackage{ifthen} 
\usepackage{lineno} 
\usepackage{makecell}
\usepackage{wrapfig}

\usepackage{algorithm}
\usepackage{algpseudocode}
\usepackage{xurl}

\usepackage{soul}
\soulregister\cite7
\soulregister\citep7
\soulregister\ref7
\soulregister\pageref7

\definecolor{mygreen}{RGB}{53,88,13}
\definecolor{myred}{RGB}{134,24,24}
\def\cmark{\color{mygreen}\ding{52}} 
\def\xmark{}
\usepackage{lineno}

\definecolor{mygray}{gray}{0.93}

\def\stretchtable{1.2}

\usepackage{siunitx}

\usepackage[authoryear]{natbib} 

\providecommand{\keywords}[1]
{
  \small	
  \textbf{\textit{Keywords---}} #1
}

\usepackage{moreverb} 

\immediate\write18{texcount -inc -incbib 
-sum main.tex > /tmp/wordcount.tex}

\immediate\write18{texcount -char -freq
 main.tex > /tmp/charcount.tex}

\begin{document}

\title{MaskSDM with Shapley values to improve flexibility, robustness, and explainability in species distribution modeling}

\author[1]{Robin Zbinden}%
\author[1]{Nina van Tiel}%
\author[1]{Gencer Sumbul}%
\author[1]{Chiara Vanalli}%
\author[2]{Benjamin Kellenberger}%
\author[1]{Devis Tuia}

\affil[1]{Environmental Computational Science and Earth Observation Laboratory, École Polytechnique Fédérale de Lausanne, Lausanne, Switzerland}%
\affil[2]{People and Nature Lab, University College London, London, United Kingdom}%
\renewcommand\Affilfont{\small}


\date{}

\begingroup
\let\center\flushleft
\let\endcenter\endflushleft
\maketitle
\endgroup



\selectlanguage{english}
\vspace{-2em}
\begin{abstract}
Species Distribution Models (SDMs) play a vital role in biodiversity research, conservation planning, and ecological niche modeling by predicting species distributions based on environmental conditions. The selection of predictors is crucial, strongly impacting both model accuracy and how well the predictions reflect ecological patterns. To ensure meaningful insights, input variables must be carefully chosen to match the study objectives and the ecological requirements of the target species. However, existing SDMs, including both traditional and deep learning-based approaches, often lack key capabilities for variable selection: (i) flexibility to choose relevant predictors at inference without retraining; (ii) robustness to handle missing predictor values without compromising accuracy; and (iii) explainability to interpret and accurately quantify each predictor's contribution. To overcome these limitations, we introduce MaskSDM, a novel deep learning-based SDM that enables flexible predictor selection by employing a masked training strategy, in which input variables are randomly hidden during training to simulate missing or ignored predictors. This approach allows the model to make predictions with arbitrary subsets of input variables while remaining robust to missing data. It also provides a clearer understanding of how adding or removing a given predictor affects model performance and predictions. Additionally, MaskSDM leverages Shapley values for precise predictor contribution assessments, improving upon traditional approximations. We evaluate MaskSDM on the global sPlotOpen dataset, modeling the distributions of 12,738 plant species. Our results show that MaskSDM outperforms imputation-based methods and approximates models trained on specific subsets of variables. These findings underscore MaskSDM's potential to increase the applicability and adoption of SDMs, laying the groundwork for developing foundation models in SDMs that can be readily applied to diverse ecological applications.

\end{abstract}%

\keywords{deep learning, explainability, flexibility, masked data modeling, robustness, shapley values, species distribution model, variable selection}

\sloppy


\section{Introduction}

In the face of the ongoing biodiversity crisis and the escalating impacts of climate change, species distribution models (SDMs) are more indispensable than ever for addressing these global challenges \citep{pollock2020protecting, portner2023overcoming}. Widely used in ecological and conservation research, SDMs are essential tools to monitor biodiversity trends \citep{jetz2019essential}, by mapping the current geographic distributions of species \citep{franklin2010mapping} and predicting their future shifts under climate change \citep{van2024regional}. Additionally, they provide critical insights into species' ecological niches, helping to inform habitat management and conservation priorities \citep{sillero2021want}. These models correlate observations of species occurrence with environmental variables \citep{elith2009species}, often focusing on abiotic factors, such as temperature, precipitation, and soil properties \citep{fourcade2018paintings}, and sometimes incorporating biotic factors, with species interactions for example \citep{wisz2013role}.
The selection of which variables to include in SDMs is critical, as the modeled outcome can vary depending on the choice of predictors \citep{araujo2006five, austin2011improving, peterson2011ecological, sillero2021want}. The input variables must align with the study objectives and the specific ecological requirements of the target species \citep{mod2016we, petitpierre2017selecting}. However, their availability is not always consistent, and traditional SDMs such as Maxent, generalized linear models (GLMs), or decision tree-based approaches \citep{valavi2022predictive}, can be affected by collinearity among predictors, particularly under collinearity shift or in cases of limited occurrence data \citep{ashcroft2011evaluation, dormann2013collinearity, feng2019collinearity}. Consequently, the number of predictors is frequently reduced, which may oversimplify the representation of the ecological processes being modeled \citep{fourcade2018paintings, cobos2019exhaustive}.

The relationships between species and their environments are inherently complex, shaped by a multitude of factors that cannot be fully captured by a limited set of variables or simplistic models. Deep learning has emerged as a promising solution to this limitation, already revolutionizing wildlife conservation and ecological research \citep{tuia2022perspectives, borowiec2022deep}. Deep learning techniques, increasingly applied to SDMs, thus referred to as DeepSDMs, leverage the vast and growing volumes of data generated by citizen science and remote sensing \citep{teng2023satbird, brun2024multispecies, picek2024geoplant, dollinger2024sat}. DeepSDMs have demonstrated remarkable capabilities, such as simultaneously mapping the global distributions of tens of thousands of species with a single model \citep{cole2023spatial}. These multi-species distribution models can identify shared patterns among species, improving predictive accuracy for those with limited occurrence data \citep{hui2013mix, cole2023spatial}. When observations are scarce, machine learning techniques such as active learning \citep{lange2023active} and few-shot learning \citep{lange2025feedforward}, combined with DeepSDMs, can be particularly effective. Unlike traditional joint species distribution models \citep{pollock2014understanding, poggiato2021interpretations}, 
DeepSDMs can also discover complex, non-linear relationships among input variables without requiring extensive predictor engineering. Moreover, such models facilitate the integration of diverse and novel data types, called \textit{modalities} in machine learning. These models can incorporate inputs such as satellite imagery or patches of rasterized predictors \citep{teng2023satbird, van2024multi}, time-series data capturing the seasonal dynamics of environmental variables \citep{picek2024geoplant}, and even textual descriptions of species ranges \citep{hamilton2024combining}. This versatility positions DeepSDMs as a powerful approach for developing generalizable, multimodal, and multi-species models that could more effectively capture underlying ecological processes. However, despite these advancements, existing approaches for SDMs (both traditional and deep learning-based) still lack critical flexibility related to the selection of predictors and face challenges in reliably interpreting variable contributions. In the following, we discuss three important limitations. 

First, SDMs should provide the \textbf{flexibility to select predictors at inference}\footnote{That is, during the prediction stage, as opposed to the training stage.} that are deemed most relevant to a specific task and target species. The applications of SDMs are numerous, each requiring a different set of predictors to be fed into the model \citep{araujo2006five, williams2012environmental, mod2016we, fourcade2018paintings}. For example, estimating the current range of a species requires incorporating human influence data along with environmental variables, as anthropogenic pressures significantly affect habitat suitability \citep{frans2024gaps}. In contrast, when modeling the potential ecological niche of a species, one may choose not to include human influence\footnote{As it is not a niche factor, though it could be incorporated separately to mitigate sampling bias.}. Similarly, while satellite imagery can offer valuable insights into the current vegetation types, its use for predicting future conditions under climate change is problematic \citep{bradley2012species}. Generally, while climatic variables are commonly used as predictors, there is less agreement about which specific additional variables to include and how these choices affect the modeled outcomes \citep{peterson2011ecological, ashcroft2011evaluation, fourcade2018paintings, williams2024predictor}. Consequently, end-users applying trained SDMs may need predictors tailored to their specific objectives, which may differ from those used during model training, thereby limiting the usability of already trained models. Moreover, existing multi-species distribution models assume the same set of input variables for all species \citep{hui2013mix}, even when the species being modeled belong to vastly different branches of the Tree of Life \citep{cole2023spatial} with varying ecological requirements \citep{williams2012environmental, petitpierre2017selecting}. Including inappropriate or non-causal predictors can result in the model learning spurious correlations, which will cause the model to fail when it is projected (e.g., spatially or temporally) in conditions where the correlation structure of predictors changes \citep{dormann2013collinearity}.
However, no existing SDMs method offers the flexibility to freely select predictors at inference, and most require substantial modifications to support this functionality. 
An alternative is to impute excluded predictors with a \textit{baseline} value \citep{ren2021can}, such as the mean of the corresponding predictor, which is assumed to have no impact on the predictions. This is questionable because the mean often corresponds to a plausible, but non-neutral, value that can inadvertently alter the predictions. 
Consequently, a common solution is to retrain new models from scratch using the desired set of predictors, following the same modeling pipeline \citep{ashcroft2011evaluation, cobos2019exhaustive}. This approach is computationally expensive and becomes increasingly impractical as the number of potential predictors grows.

\begin{figure}[tb]
    \centering
    \includegraphics[width=1\linewidth]{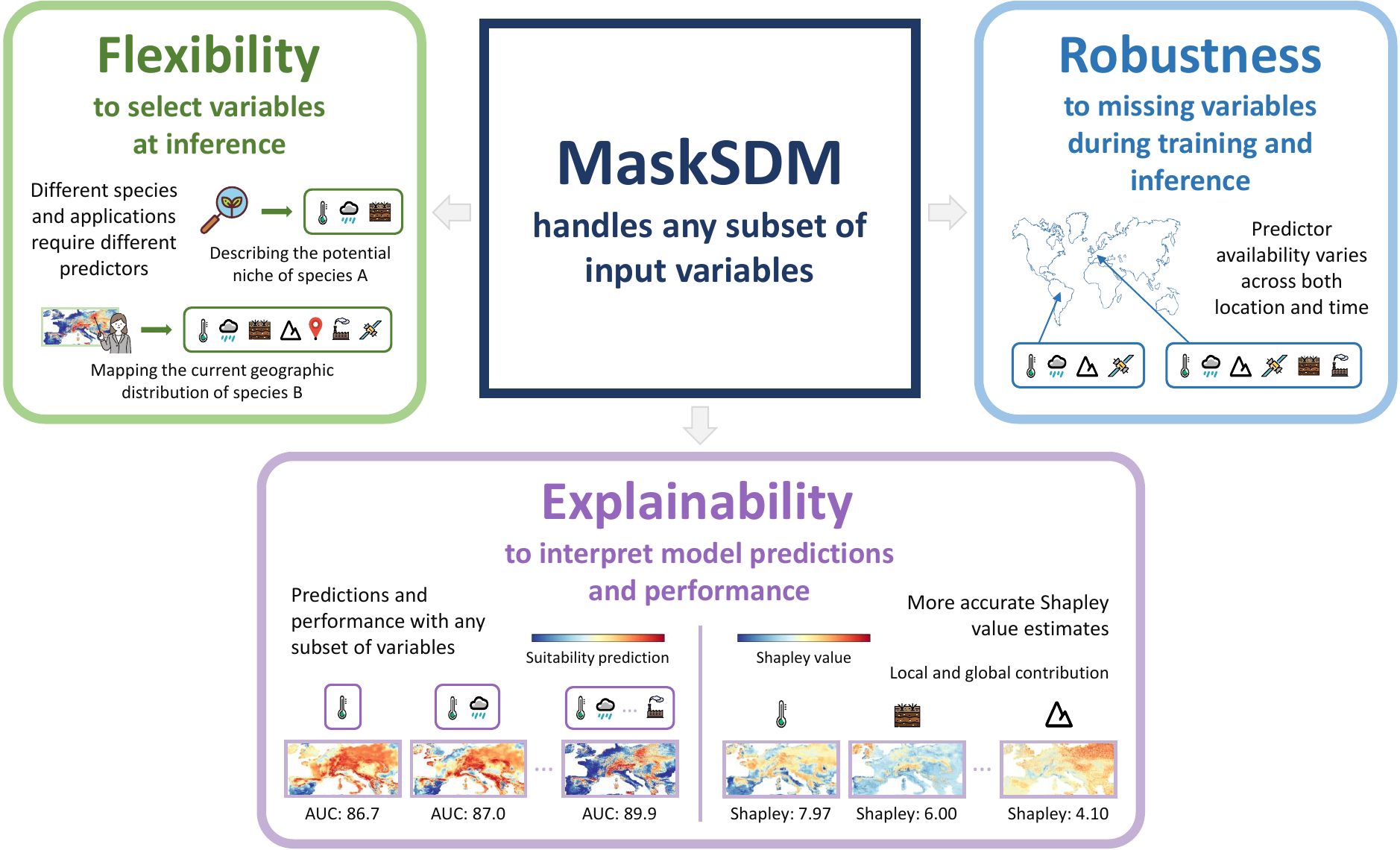}
    \caption{Overview of the capabilities of the proposed MaskSDM approach. MaskSDM can process any subset of predictors, allowing a flexible selection of variables at inference to adapt to specific study questions and species. It is robust to missing data and enhances model explainability, particularly by providing more accurate Shapley value estimates.}
    \label{fig:use_overview}
\end{figure}

A second important requirement for SDMs is \textbf{robustness to missing predictor values}, both during training and inference. Geospatial predictors used by SDMs are often inconsistently available across the globe \citep{bucklin2015comparing}. These predictors are typically derived from rasters generated by predictive models, such as WorldClim \citep{hijmans2005very} or SoilGrids \citep{hengl2017soilgrids250m}. However, the outputs of these models can be highly inconsistent and noisy, particularly in regions that have been sparsely sampled during their development. For instance, the precision of WorldClim variables deteriorates in areas with few weather stations and steep climatic gradients, such as regions with high variation in elevation \citep{hijmans2005very}. Additionally, some rasters may lack complete coverage of the areas of interest. For example, some digital elevation maps exclude high-latitude regions \citep{farr2007shuttle}, limiting the direct applicability of SDMs to those areas. Similar limitations affect remote sensing products, where data acquisition can be hindered by factors such as cloud cover \citep{gerber2018predicting}. Another challenge arises when predictor availability differs between training and inference.
For example, a model trained in a region with rich environmental data, such as Switzerland \citep{kulling2024sweco25}, may incorporate locally specific predictors that are unavailable elsewhere \citep{guisan2025spatially}. This can hinder the transferability of the model, as most SDMs assume that the same set of predictors is available during both training and prediction.
While missing values can sometimes be replaced or reconstructed using interpolation between neighboring values \citep{kornelsen2014comparison}, this approach requires additional data close to the location of interest, which is not always available or easy to obtain.

Finally, an essential feature expected of SDMs is their \textbf{explainability}. In most SDM applications, the goal extends beyond obtaining a single suitability score; understanding the factors that drive the prediction of the model is equally important \citep{ashcroft2011evaluation, barbet201440, ryo2021explainable}. This is particularly critical when communicating model outputs to policymakers for conservation decisions, where transparency and interpretability are essential \citep{guisan2013predicting}. Importantly, gaining insights into the contributions of different predictors can shed light on the underlying ecological processes \citep{ryo2021explainable}. While explanations derived from correlative SDMs are inherently limited in their causal power and must be interpreted cautiously, they can still reveal patterns and provide a better mechanistic understanding of the factors that define suitable habitats for a species. 
Ideally, SDMs should not only assess the impact of adding new predictors to a given set but also provide a single value per predictor representing its average contribution.
Some traditional SDMs methods allow for analyzing the contributions of different input variables. For example, linear regression models provide direct access to variable weights, but caution is needed when interpreting these weights in the presence of collinear predictors \citep{dormann2013collinearity}. Decision trees can also help reveal predictor contributions by highlighting which variables are used for splits. Deep learning methods, on the other hand, are notoriously difficult to interpret. To understand the impact of individual variables or groups of variables in these models, researchers typically perform ablation studies \citep{cole2023spatial, dollinger2024sat, picek2024geoplant}, which involve training multiple models with different subsets of variables, a computationally expensive process. To address these challenges, the field of \textit{eXplainable Artificial Intelligence} (XAI) has developed methods to make machine learning models more interpretable \citep{ribeiro2016should, gunning2019xai}. One popular approach is based on the computation of Shapley values \citep{shapley1953value}, which summarize the average contribution of a variable or group of variables \citep{lundberg2017unified, covert2020understanding}. Shapley values have several desirable properties, including greater robustness to correlated predictors compared to alternative approaches such as Jackknife metrics, and are increasingly used in SDMs \citep{cha2021interpretable, maloney2022explainable, bourhis2023explainable}. However, their computation requires a model capable of handling flexible inputs, i.e., able to accept any subset of predictors, which is generally not possible with current SDMs.  As a result, Shapley values are typically approximated \citep{lundberg2017unified}. These approximations rely on strong assumptions, such as model linearity or predictor independence, both of which are rarely satisfied in SDMs, where models are frequently non-linear and predictors tend to be strongly correlated \citep{dormann2013collinearity, aas2021explaining}. This limitation highlights the need for more accurate methods to compute Shapley values.

All these properties, i.e., flexibility, robustness, and explainability, can be achieved if a model is capable of considering any subset of variables at any time while still making accurate predictions based on the available data. To this end, we introduce MaskSDM, a novel deep learning method that achieves this by modifying the training process to randomly mask certain input variables \citep{devlin2018bert, majmundar2022met, du2023remasker, mizrahi20244m}. In doing so, our approach trains the model to make predictions using only a reduced set of predictors. Although the random masking procedure does not encompass all possible subsets of predictors, it effectively explores the predictor space, enabling the model to accommodate missing values using a specialized \textit{token} to indicate the absence of the considered predictors. This ensures accurate predictions even when some variables are unavailable, directly addressing the issue of missing predictors. 
This provides end-users with the flexibility to select variables they consider relevant for their particular application or species of interest. Furthermore, MaskSDM facilitates a deeper understanding of predictor roles by allowing users to analyze predictions and performance based on specific subsets of variables.

In this paper, we demonstrate the potential of MaskSDM in several ways. First, we compare MaskSDM to several alternative baselines, including imputing methods and an “oracle" method that requires training a separate model for each subset of variables. Our results show that MaskSDM outperforms all imputing approaches and approximates the predictions and performance of the oracle method. Second, we conduct analyses on the performance and predictions for different subsets of variables on a large set of species. Third, we illustrate how MaskSDM integrates seamlessly with Shapley values to explain model predictions and quantify individual predictor contributions. Unlike traditional SDMs employing Shapley values, MaskSDM does not rely on strong assumptions of predictor independence. Using this approach, we produce maps of Shapley values that highlight regions where specific predictors play more prominent roles. All experiments are conducted on the open-access, global sPlotOpen dataset \citep{sabatini2021splotopen}, which consists of presence-absence plant observations in plots. This enables the creation of global prediction maps along with associated predictor contributions for the \num{12738} species considered.
The capabilities of MaskSDM are summarized in Fig. \ref{fig:use_overview}. The code is available at \url{https://github.com/zbirobin/MaskSDM}.

Our findings highlight the advantages of MaskSDM in advancing ecological research by providing researchers with greater flexibility to formulate and test ecological hypotheses. This approach also sets the stage for the development of a \textit{foundation model} in SDMs \citep{bommasani2021opportunities}, which could leverage the integration of a large number of relevant predictors combined with extensive observations spanning multiple species. Such a generic model could be readily adapted to meet the specific needs of its users, enhancing its utility across diverse SDMs applications. In Appendix \ref{appendix:workflow}, we present a workflow illustrating how MaskSDM can be applied across different scenarios.

\section{Material and Methods}

In this section, we: i) present the MaskSDM method and describe how it overcomes critical limitations of traditional SDMs in Section \ref{sec:masksdm}; ii) explain how MaskSDM can be leveraged to improve estimates of Shapley values in Section \ref{sec:shapley}; and iii) outline the experimental setup used in this study to evaluate our approach in Section \ref{sec:setup}, since MaskSDM is a general framework with multiple possible implementations.

\subsection{MaskSDM}
\label{sec:masksdm}

Traditional SDMs aim to predict the likelihood of observing a species in a given location, based on a predefined and fixed set of input variables, denoted as $F = \{x_1, x_2, \ldots, x_M\}$ \citep{valavi2022predictive}. 
Each predictor $x_i$ represents an environmental variable or another factor hypothesized to influence species distributions. They can encompass various data types, including commonly used tabular data \citep{valavi2022predictive}, but also more complex data types, such as satellite imagery \citep{gillespie2024deep, dollinger2024sat}, climatic time series \citep{picek2024geoplant}, or even textual descriptions of the location \citep{cheng2023teaw}.
SDMs are trained to relate these predictors $F$ to sparse species occurrence data. 

\begin{figure}
    \centering
    \includegraphics[width=1\linewidth]{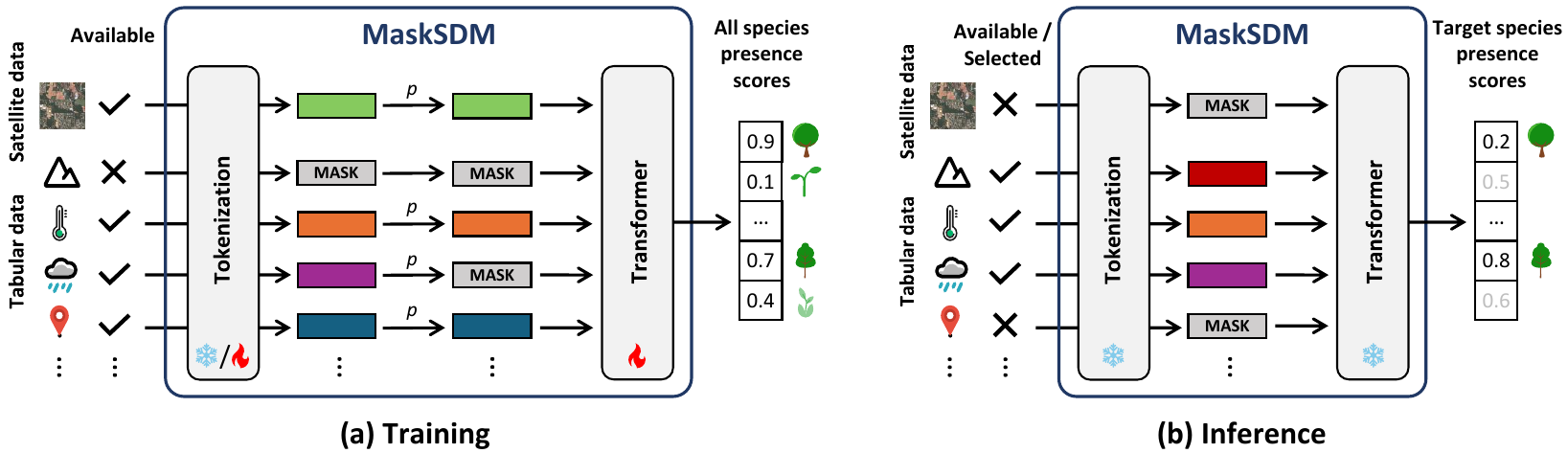}
    \caption{Overview of MaskSDM. (a) During training, our method employs a mask token to indicate missing input variables to the Transformer model. Additionally, this mask token is used to randomly mask each input variable with a probability $p$. (b) During inference, MaskSDM can take any subset of variables as input to predict the presence of species of interest.}
    \label{fig:architecture_overview}
\end{figure}

To address the limitations associated with a fixed set of predictors, we propose a novel method designed to predict the presence of species using any subset $S \subseteq F$ of variables that are available at the given location and are deemed relevant for the particular species and application of interest. MaskSDM leverages masked data modeling to learn species distributions in a supervised manner. This is done by randomly masking input variables during training, forcing the model to learn the distribution despite missing variables. This approach enables the model to adaptively handle varying subsets of predictors during both training and inference. The overall approach of MaskSDM is illustrated in Fig. \ref{fig:architecture_overview} and described in detail below.

\subsubsection{Tokenization}
The input variables are first converted into a standardized format through a process called \textit{tokenization}, which involves projecting inputs of heterogeneous types into high-dimensional feature vectors (\textit{tokens}) of predefined size \citep{gorishniy2021revisiting, gorishniy2022embeddings, mizrahi20244m}.
The functions $g$ that produce these tokens, defined as $t_i = g_i(x_i) \in \mathbb{R}^d$ for each input variable $x_i \in F$, are known as \textit{tokenizers}. Each predictor $x_i$ has a dedicated tokenizer $g_i$, operating solely on its corresponding predictor. As a result, removing or replacing the associated token $t_i$ eliminates the information in $x_i$, enabling the selective omission of specific variables from the model.
For example, in tabular data, each variable may be tokenized independently, allowing individual variables to be excluded if necessary. The tokenizers' parameters are trained alongside the rest of the model parameters. Additionally, for certain data types, such as satellite images, pre-trained tokenizers can be leveraged to generate more informative and general tokens, while also reducing computational costs \citep{klemmer2023satclip, mizrahi20244m}.

\subsubsection{Transformer}

After tokenization, we employ a deep learning model called a \textit{transformer encoder} \citep{vaswani2017attention}, which is designed to capture complex interactions among input variables. The transformer encoder takes the tokens as inputs and predicts a presence score for each species, learning relationships and interactions between tokens through a mechanism known as self-attention \citep{lin2017a}. By considering these token interactions, the transformer encoder can account for non-linear combinations of environmental factors.

\subsubsection{Masked Data Modeling}
During training, MaskSDM utilizes the masked modeling paradigm to learn robust species distributions. Masked data modeling is a deep learning approach originally developed in natural language processing \citep{devlin2018bert} and later adapted for computer vision \citep{he2022masked}, serving as a task to help models learn more meaningful representations of data. This method involves \textit{masking}, or hiding, a portion of the input data and training the model to reconstruct the missing part. For instance, in Masked Language Modeling (MLM), a text model is provided with a sentence in which certain words are hidden and replaced by a \textit{mask} token. The model is then tasked with predicting the missing words, which encourages it to learn the underlying structure and semantics of language in an unsupervised manner. 

In MaskSDM, we adapt this approach for supervised learning in species distribution modeling. Similarly to MLM, we replace missing input variables with a learned \textit{mask token} $t_{\text{MASK}}$, which signals to the transformer encoder that a predictor is absent. This mask token is learned as part of the model training process, alongside the parameters of the tokenizers and transformer encoder. By incorporating mask tokens, we can leverage all available samples during training, even if some values are missing.

To enhance the model's robustness to varying subsets of input variables, we randomly mask additional input variables during training, even when they are available \citep{mizrahi20244m}. At each training iteration, the probability of masking each input variable $p$ is drawn uniformly between \num{0} and \num{1}. The tokens corresponding to the masked input variables are then replaced with the mask token $t_{\text{MASK}}$. This stochastic masking strategy helps the model to effectively handle scenarios where only a limited subset of variables is accessible for predicting species distributions, while also handling cases where nearly all variables are available. During inference, MaskSDM enables the replacement of missing, unsuitable, or irrelevant variables with the mask token, ensuring flexibility in model predictions. It also allows users to test different subsets of variables, revealing their impact on prediction maps and model performance. 

\subsection{Shapley values with MaskSDM}
\label{sec:shapley}

MaskSDM enables analyzing prediction and performance changes across variable subsets without training separate models, as it can naturally handle any combination of variables.
However, since the number of subsets grows exponentially ($2^M$ subsets for $M$ predictors), exhaustive evaluation is impractical. 
Therefore, we use Shapley values to measure predictor importance, taking into account its average contribution to model predictions or performance \citep{shapley1953value, lundberg2017unified}. The Shapley value $\phi_i$ for variable $x_i$ represents its average contribution across subsets of variables and is defined as:
\begin{equation} 
    \phi_i = \sum_{S \subseteq F \setminus \{x_i\}} \frac{|S|(|F| - |S| - 1)!}{|F|!}\left[f(S \cup \{x_i\}) - f(S)\right],
\label{eq:shapley}
\end{equation}
where $f$ denotes the model output or performance metric. To compute Shapley values, $f$ must be able to consider a subset of predictors only. For models that cannot inherently handle subsets, approximations are often used, typically assuming predictor independence or model linearity—assumptions that are often violated in SDMs \citep{dormann2013collinearity}. These approximations usually replace excluded variables with \textit{baseline} values, such as their mean, which can significantly bias predictions \citep{lundberg2017unified, ren2021can}.
MaskSDM overcomes this limitation by enabling predictions based directly on subsets of variables, providing more reliable Shapley value estimates and making it a robust tool for assessing predictor importance in SDMs.

Another common challenge in computing Shapley values is the exponential number of terms in the sum to calculate, one for each subset. When the number of predictors is large, this summation is typically approximated using Monte Carlo methods, which involve randomly sampling $k$ of these terms. As $k$ approaches $2^M$, the estimate converges to the true Shapley value. However, we observe that convergence can be very slow, as it strongly depends on the subsets $S$ considered. Specifically, adding any variable to the empty set $\{\emptyset \}$ significantly alters the predictions and considerably improves performance. Consequently, the estimate of the Shapley value becomes heavily influenced by how frequently the empty set is sampled. To address this, we ensure that the subsets considered are evenly distributed across all sizes and leverage \textit{Latin squares} \citep{keedwell2015latin} to improve computational efficiency. The exact procedure is detailed in Appendix \ref{sec:appendix_shapley} and is referred to as the \textit{stratified} Monte Carlo method, distinguishing it from the \textit{uniform} Monte Carlo approach, which selects subsets fully at random. This stratified Monte Carlo approach allows us to compute Shapley values efficiently, which is crucial given our case with \num{61} predictors. When variables are grouped, i.e., when $x_i$ in Equation \ref{eq:shapley} represents a group of predictors rather than individual variables (e.g., \num{6} groups), we compute the exact Shapley values since the computation becomes much faster.

\subsection{Experimental setup}
\label{sec:setup}

\subsubsection{Dataset}

We use the sPlotOpen dataset \citep{sabatini2021splotopen}, which includes \num{95104} vegetation plots worldwide. We retain plant species with more than 20 recorded presences, resulting in \num{12738} species. We employ spatial block cross-validation \citep{roberts2017cross} to split the data into training, validation, and test sets (see Appendix \ref{sec:appendix_dataset} for visualization of their geographic distribution).
Each spatial block spans an area of 1°$\times$1°, and the blocks are randomly assigned to the splits while maintaining a \num{70}:\num{15}:\num{15} ratio for training, validation, and testing, respectively. The validation set is used to optimize hyperparameters and apply early stopping. We evaluate the model on the test set, considering only species with at least one observation in each split (\num{9009} species). As a case study, we select three European plant species for a qualitative analysis of their predictions. These species are \textit{Anthyllis vulneraria}, a medicinal plant native to Europe, also known as kidney vetch; \textit{Vaccinium myrtillus}, a small deciduous shrub, also referred to as European blueberry or bilberry; and \textit{Quercus ilex}, commonly known as the holm oak, a large evergreen tree.

\begin{table}
\centering
\renewcommand{\arraystretch}{0.72}
\setlength\tabcolsep{4pt}
\small
\begin{tabular}{@{}llllll@{}}
\toprule
\textbf{Group} & \textbf{Shapley} & \textbf{Variable}    & \textbf{Description} & \textbf{\%Missing}   & \textbf{Shapley} \\ \midrule
WorldClim      & 9.31 & bio\_1                    & Annual mean temperature & 0 & 1.63           \\
               &  & bio\_2                    & Mean diurnal range & 0 & 1.30            \\
               &  & bio\_3                    & Isothermality & 0 & 1.71            \\
               &  & bio\_4                    & Temperature seasonality & 0 & 1.72            \\
               &  & bio\_5                    & Max temperature of warmest month & 0 & 1.51            \\
               &  & bio\_6                    & Min temperature of coldest month & 0 & 1.76            \\
               &  & bio\_7                    & Temperature annual range & 0 & 1.59            \\
               &  & bio\_8                    & Mean temperature of wettest quarter & 0 & 1.41            \\
               &  & bio\_9                    & Mean temperature of driest quarter & 0 & 1.42            \\
               &  & bio\_10                    & Mean temperature of warmest quarter & 0 & 1.48            \\
               &  & bio\_11                    & Mean temperature of coldest quarter & 0 & 1.79            \\
               &  & bio\_12                    & Annual precipitation & 0 & 1.30            \\
               &  & bio\_13                    & Precipitation of wettest month & 0 & 1.26            \\
               &  & bio\_14                    & Precipitation of driest month & 0 & 1.22            \\
               &  & bio\_15                    & Precipitation seasonality & 0 & 1.00            \\
               &  & bio\_16                    & Precipitation of wettest quarter & 0 & 1.34            \\
               &  & bio\_17                    & Precipitation of driest quarter & 0 & 1.24            \\
               &  & bio\_18                    & Precipitation of warmest quarter & 0 & 1.33            \\
               &  & bio\_19                    & Precipitation of coldest quarter & 0 & 1.12            \\\midrule
SoilGrids      & 8.18 & ORCDRC                    & Soil organic carbon content & 0 & 1.06            \\
               &  & PHIHOX                    & pH index measured in water solution & 0 & 1.23            \\
               &  & CECSOL                    & Cation Exchange Capacity of soil & 0 & 0.75            \\
               &  & BDTICM                    & Absolute depth to bedrock & 0 & 0.64            \\
               &  & CLYPPT                    & Weight percentage of the clay particles & 0 & 0.83            \\
               &  & SLTPPT                    & Weight percentage of the silt particles & 0 & 1.17            \\
               &  & SNDPPT                    & Weight percentage of the sand particles & 0 & 0.89            \\
               &  & BLDFIE                    & Bulk density & 0 & 1.11            \\\midrule
Topography     & 4.73 & Elevation                 & Elevation & 7.50 & 1.06            \\
               &  & Aspect                    & Aspect & 8.12 & 0.00            \\
               &  & Slope                     & Slope & 8.12  & 0.72            \\\midrule
Location       & 8.90  & Longitude                 & Longitude & 0 & 2.08            \\
               &  & Latitude                  & Latitude & 0 & 1.93            \\\midrule
Human Inf.     & 4.45 & HFP2009                  & Human footprint & 0 & 0.46          \\
               &  & Built2009                 & Built environments & 0 & 0.07          \\
               &  & Croplands2005             & Crop lands & 0 & 0.20          \\
               &  & Lights2009                & Nightlights & 0 & 0.31          \\
               &  & Navwater2009              & Navigable waterways & 0 & 0.22          \\
               &  & Pasture2009               & Pasture lands & 0 & 0.32          \\
               &  & Popdensity2010            & Population density & 0 & 0.80          \\
               &  & Railways                  & Railways & 0 & 0.03          \\
               &  & Roads                     & Major roadways & 0 & 0.21            \\\midrule
Metadata       & 7.02 & Releve\_area          & Surface area & 0.03 & 1.20            \\
               &  & Location\_uncertainty     & Location uncertainty & 29.53 & 0.65            \\
               &  & Cover\_total              & Total cover & 79.59 & 0.37            \\
               &  & Cover\_tree\_layer        & Tree layer cover & 87.28 & 0.47            \\
               &  & Cover\_shrub\_layer       & Shrub layer cover & 82.33 & 0.42            \\
               &  & Cover\_herb\_layer        & Herb layer cover & 68.80 & 0.58            \\
               &  & Cover\_moss\_layer        & Moss layer cover & 89.82 & 0.24            \\
               &  & Cover\_lichen\_layer      & Lichen layer cover & 99.26 & 0.00            \\
               &  & Cover\_algae\_layer       & Algae layer cover & 99.96 & 0.00            \\
               &  & Cover\_litter\_layer      & Litter layer cover & 96.68 & 0.05            \\
               &  & Cover\_bare\_rocks        & Bare rocks cover & 97.11 & 0.10            \\
               &  & Cover\_cryptogams         & Cryptogams cover & 99.19 & 0.03            \\
               &  & Cover\_bare\_soil         & Bare soil cover & 97.11 & 0.06            \\
               &  & Height\_trees\_highest    & Height of tallest trees & 91.36 & 0.42            \\
               &  & Height\_trees\_lowest     & Height of shortest trees & 99.53 & 0.01            \\
               &  & Height\_shrubs\_highest   & Height of tallest shrubs & 96.44 & 0.10            \\
               &  & Height\_shrubs\_lowest    & Height of shortest shrubs & 99.72 & 0.01            \\
               &  & Height\_herbs\_average    & Average height of herbs & 93.80 & 0.12            \\
               &  & Height\_herbs\_lowest     & Height of shortest herbs & 99.48 & 0.00            \\
               &  & Height\_herbs\_highest    & Height of tallest herbs & 98.86 & 0.03              \\ \hline \midrule
Sum            &  48.0 & Sum                      & &  & 48.0             \\\bottomrule
\end{tabular}

\caption{All predictors used, ordered by groups, with associated Shapley values and number of missing values. The sum of the Shapley values is equal to 42.6.}
\label{tab:predictors}
\end{table}

We gather predictor variables from various sources for each vegetation plot in the dataset. Climate data, including temperature and precipitation statistics at a resolution of 1 km², are obtained from WorldClim \citep{hijmans2005very}. Soil properties relevant to plant species, such as organic carbon content, pH levels, and texture, are sourced at a resolution of 250 meters from SoilGrids \citep{hengl2017soilgrids250m}. 
We also include topographic information—elevation, slope, and aspect—derived from the 90-meter resolution Shuttle Radar Topography Mission digital elevation model \citep{farr2007shuttle}, version 4. Additionally, we integrate human influence data from human footprint maps, which include nine variables such as population density and nightlight intensity \citep{venter2016global}, available at a 1 km² resolution.
The longitude and latitude coordinates are also provided to the model, as spatial information has been shown to enhance SDM performance, especially in contexts where geographic factors play a significant role in species distributions \citep{elith2009species, domisch2019spatially}.
The sPlotOpen dataset also includes supplementary metadata for some plots, such as location uncertainty, plot surface area, and vegetation layer coverage and height. These metadata, while sometimes incomplete, can be highly predictive of species distributions and help disentangle variable contributions during model training. Altogether, these sources yield 61 tabular predictor variables, all standardized before being inserted into the model. Finally, image features derived from Sentinel-2 satellite images are incorporated using SatCLIP, a pretrained location encoder that maps geographic coordinates to compact feature vectors based on the surrounding satellite imagery \citep{klemmer2023satclip}. These features capture environmental context and are used in addition to the 61 tabular predictors. While WorldClim, SoilGrids, SatCLIP, and coordinate variables are consistently available for all plots, other variables may sometimes be missing. An exhaustive list of all predictors is provided in Table \ref{tab:predictors}.

\subsubsection{Model architecture and training}

Tabular inputs are tokenized using periodic activation functions \citep{gorishniy2022embeddings}, while satellite image tokens are obtained using the SatCLIP encoder \citep{klemmer2023satclip}. The transformer encoder follows the FTTransformer architecture \citep{gorishniy2021revisiting}, and outputs suitability scores for the \num{12738} species. Model parameters are learned using the schedule-free AdamW optimizer \citep{loshchilov2017decoupled, defazio2024road} with weighted binary cross-entropy \citep{zbinden2024selection}. Further details on the model architecture and training procedure can be found in Appendix \ref{sec:architecture_details}.

\subsubsection{Baselines}

We include several baselines to analyze the effectiveness of MaskSDM in building a model that considers only a subset of predictors as input. We first establish an upper bound on the model's achievable performance, referred to as the \textit{oracle}. The oracle is constructed by training separate models for each considered subset of predictors, with each model trained and evaluated exclusively on its respective subset of variables.
This approach is computationally expensive because capturing all possible combinations of predictors would require training $2^M$ models, where $M$ is the total number of predictors. Consequently, we limit the comparison to a subset of predictor combinations and use it to assess the performance gap between the oracle and MaskSDM.

We then compare MaskSDM to four imputation-based methods for handling missing variables, using the following neutral baseline values: \textit{mean}, \textit{median}, \textit{marginal}, and \textit{conditional}. Detailed implementation descriptions are provided in Appendix \ref{sec:baseline_details}. Ideally, these baseline values should not affect the prediction, ensuring the model genuinely considers only a subset of predictors as input. The drawbacks of both the marginal and conditional imputation methods are that they require access to the training set during inference and increase inference time.

\section{Results}

\subsection{Comparison of MaskSDM with baselines}

\begin{table}
    \renewcommand{\arraystretch}{\stretchtable}
    \setlength{\arrayrulewidth}{0.2pt}
    \setlength\tabcolsep{6pt}
    \centering
    \small
    \begin{tabular}
        {|l|l|cccccccccc|}
        \hline
        \multirow{8}{*}{\rotatebox[origin=c]{90}{\textbf{Predictors (\#)}}} 
        & Avg. Temperature (1) & \cmark & \xmark & \xmark & \cmark & \cmark & \cmark & \cmark & \cmark & \cmark & \cmark \\
        & \cellcolor{mygray}WorldClim (19) & \cellcolor{mygray}\xmark & \cellcolor{mygray}\xmark & \cellcolor{mygray}\xmark & \cellcolor{mygray}\cmark & \cellcolor{mygray}\cmark & \cellcolor{mygray}\cmark & \cellcolor{mygray}\cmark & \cellcolor{mygray}\cmark & \cellcolor{mygray}\cmark  & \cellcolor{mygray}\cmark \\
        & SoilGrids (8) & \xmark & \xmark & \xmark & \xmark & \cmark & \cmark & \cmark & \cmark & \cmark  & \cmark \\
        & \cellcolor{mygray}Topographic (3) & \cellcolor{mygray}\xmark & \cellcolor{mygray}\xmark & \cellcolor{mygray}\xmark & \cellcolor{mygray}\xmark & \cellcolor{mygray}\xmark & \cellcolor{mygray}\cmark & \cellcolor{mygray}\cmark & \cellcolor{mygray}\cmark &  \cellcolor{mygray}\cmark  & \cellcolor{mygray}\cmark\\
        & Location (2) & \xmark & \cmark & \xmark & \xmark & \xmark & \xmark & \cmark & \cmark & \cmark  & \cmark\\
        & \cellcolor{mygray}Human footprint (9)  & \cellcolor{mygray}\xmark & \cellcolor{mygray}\xmark & \cellcolor{mygray}\xmark & \cellcolor{mygray} \xmark & \cellcolor{mygray}\xmark & \cellcolor{mygray}\xmark & \cellcolor{mygray}\xmark & \cellcolor{mygray}\cmark & \cellcolor{mygray}\cmark  & \cellcolor{mygray}\cmark \\
        & Plot metadata (20) & \xmark & \xmark & \xmark & \xmark & \xmark & \xmark & \xmark & \xmark & \cmark  & \cmark \\
        & \cellcolor{mygray}Satellite image features & \cellcolor{mygray}\xmark & \cellcolor{mygray}\xmark & \cellcolor{mygray}\cmark & \cellcolor{mygray}\xmark & \cellcolor{mygray}\xmark & \cellcolor{mygray}\xmark & \cellcolor{mygray}\xmark & \cellcolor{mygray}\xmark & \cellcolor{mygray}\xmark  & \cellcolor{mygray}\cmark \\
         \hline
        \multirow{9}{*}{\rotatebox[origin=c]{90}{\textbf{Method}}}
        & \textbf{Imputing}:        &  &  &  &  &  &  &  &  &  &  \\
        & Mean              & 69.6 & 77.8 & 75.6 & 88.7 & 91.3 & 91.8 & 95.5 & 95.5 & 95.9 & 97.8 \\
        & Median            & 73.5 & 82.4 & 74.7 & 89.4 & 91.1 & 91.6 & 95.9 & 95.8 & 96.3 & \underline{97.9} \\
        & Marginal          & 62.4 & 79.8 & 82.1 & 89.5 & 92.7 & 93.2 & 96.1 & 96.1 & 96.6 & 97.8 \\ 
        & Conditional       & 72.6 & \underline{96.6} & 95.0 & 96.6 & 97.0 & 97.0 & 97.2 & 97.0 & 97.6 & 97.8 \\\cline{2-12}
        & \textbf{Masking}:        &  &  &  &  &  &  &  &  &  &  \\
        & MaskSDM (ours)    & \underline{85.1} & 96.0 & \underline{95.7} & \textbf{97.1} & \textbf{97.4} & \textbf{97.5} & \underline{97.6} & \textbf{97.6} & \textbf{98.0} & \textbf{98.0} \\ \cline{2-12}
        & \textbf{Oracle}:        &  &  &  &  &  &  &  &  &  &  \\
        & One model per column & \textbf{86.3} & \textbf{96.8} & \textbf{96.7} & \textbf{97.1} & \textbf{97.4} & \textbf{97.5} & \textbf{97.7} & \textbf{97.6} & \underline{97.9} & 97.8  \\ \hline
    \end{tabular}%
    \caption{Mean test AUC performance comparison of MaskSDM to the baselines across subsets of input variables. Each column represents a different subset of predictors used. For MaskSDM and imputing baselines, a single model produces the entire row of results, while the oracle baseline requires training a separate model for each column. Bold values indicate the best performance per column, and underlined values represent the second-best score. Numbers in parentheses indicate the number of input variables in each subset. Note that the average temperature is included in the WorldClim data. The other subsets do not overlap.}
    \label{tab:baselines}
\end{table}


We compare the performance of MaskSDM to the baselines for different subsets of predictors, using the mean AUC\footnote{Area Under the receiver operating characteristic Curve, expressed as a percentage for readability} across all the species computed on the test set (Table \ref{tab:baselines}). MaskSDM performs as well as the oracle when at least two predictors are considered, effectively approaching the best possible outcome a model can achieve for a given subset of predictors. However, while the oracle approach requires training a separate model for each predictor subset (one model per column in Table \ref{tab:baselines}), MaskSDM and the imputing baselines require training a single model to generate their row of results. When fewer predictors are used, there is a small performance gap between MaskSDM and the oracle that can be reduced by extending the training time of MaskSDM (Table \ref{tab:epoch} in Appendix \ref{sec:appendix_epoch}). Interestingly, when many predictors are included, MaskSDM outperforms the oracle. We hypothesize that this occurs because the missing values in the training and test sets are imputed in the oracle baseline. Such an operation can introduce artifacts that degrade model performance, whereas the MaskSDM learning strategy avoids imputation.

We find that MaskSDM consistently outperforms all other imputing baselines by a significant margin, especially when fewer predictors are used. These large performance differences are evident even before the model has fully converged during training (Table \ref{tab:epoch} in Appendix \ref{sec:appendix_epoch}). Specifically, MaskSDM achieves its maximum validation AUC at 178 epochs, yet its performance on the test set is already nearly optimal as early as at 25 epochs. This indicates that MaskSDM can achieve high performance with minimal additional training. Moreover, the distribution maps predicted by MaskSDM align more closely with those of the oracle model compared to the imputation baselines (Table \ref{tab:diff_oracle} and Fig. \ref{fig:oracle_maps_comparison} in Appendix \ref{sec:appendix_results}). Finally, we evaluate MaskSDM against alternative modeling approaches, including Maxent \citep{maxent} and architectures such as ResNet \citep{gorishniy2021revisiting}, using both the sPlotOpen dataset and a second dataset, GeoPlant \citep{picek2024geoplant}, which consists of vegetation plots in Europe. Due to space constraints, the results are presented in Appendix \ref{sec:appendix_results}. As shown in Tables \ref{tab:architecture} and \ref{tab:geoplant}, MaskSDM consistently outperforms the other approaches for both datasets.

\subsection{Predictor impact on performance}

We examine the performance of MaskSDM across the subsets of predictors (Table \ref{tab:baselines}). Notably, using only WorldClim variables already yields high performance. Additional improvements can be achieved by incorporating SoilGrids, location data, and metadata. Metadata, in particular, contributes to an increase of \num{0.4}\% in performance. Although metadata is often unavailable, these results suggest that including it when possible can significantly enhance predictions. Otherwise, the differences in performance between subsets are relatively small. This is likely due to the limited sample size of presence data for most species, which diminishes the benefits of adding more predictors. We show that the impact of using different subsets of variables depends on the number of species presence observations available (Table \ref{tab:num_occ} in Appendix \ref{sec:appendix_occ}). Specifically, greater performance differences are observed when species have more presence records, consistent with classical statistical theory. 

\begin{figure}
    \centering
    \includegraphics[width=1\linewidth]{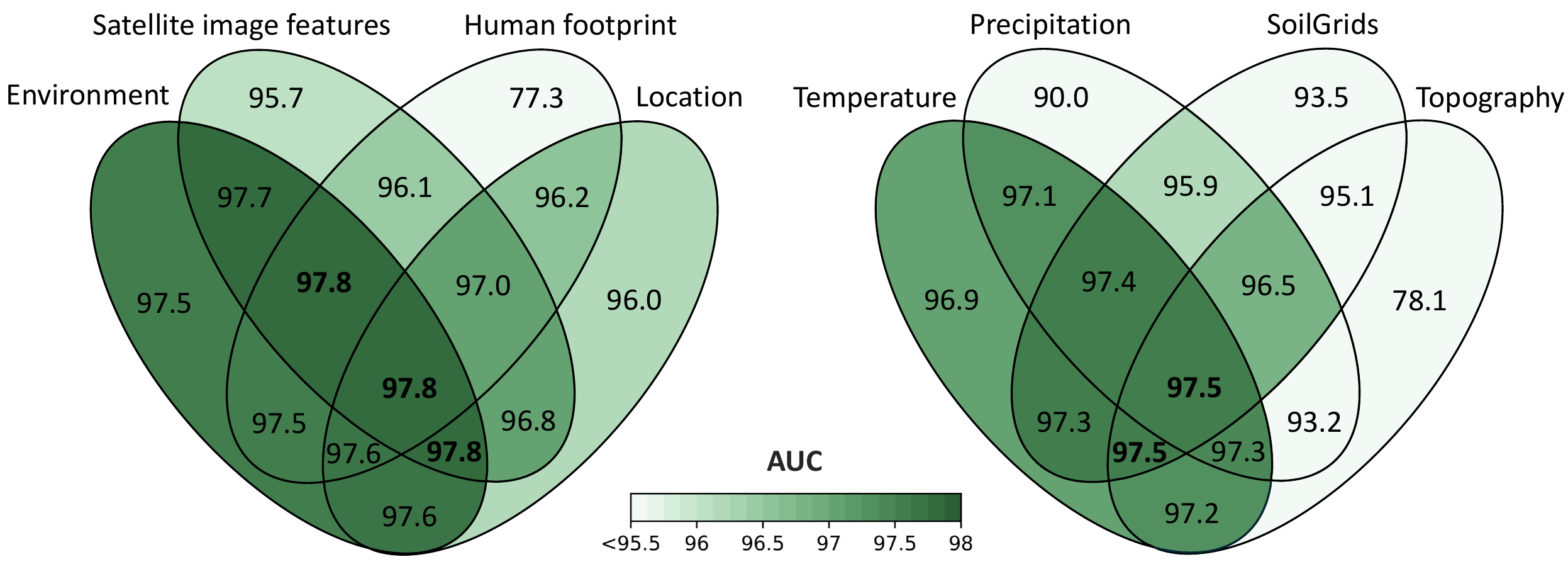}
    \caption{Mean AUC performance on the test set for different subsets of predictors using MaskSDM. Each ellipse represents a group of variables, and their intersection indicates the AUC performance when the union of the corresponding variables is used as input to the model. The bold numbers highlight the subset that maximizes performance within each Venn diagram. \textbf{Left:} Four groups of predictors, where the \textit{Environment} predictors group includes predictors from WorldClim (temperature and precipitation), SoilGrids, and topographic information. \textbf{Right:} The predictor groups that make up the \textit{Environment} group.}
    \label{fig:venn_diagram}
\end{figure}

Further decomposing which combinations of predictors lead to optimal performance, we observe that combining the \textit{environment} group (which includes WorldClim, SoilGrids, and topography variables) with satellite image features results in the best performance using the fewest predictors (Fig. \ref{fig:venn_diagram}, left panel). The environment group alone also produces strong performance. Within the environmental predictor group, we find that temperature variables from WorldClim play a crucial role (Fig. \ref{fig:venn_diagram}, right panel). Importantly, considering all four groups of variables seems essential to achieve optimal performance.

\subsection{Prediction maps of species occurrence}

\begin{figure}
    \centering
    \includegraphics[width=0.765\linewidth]{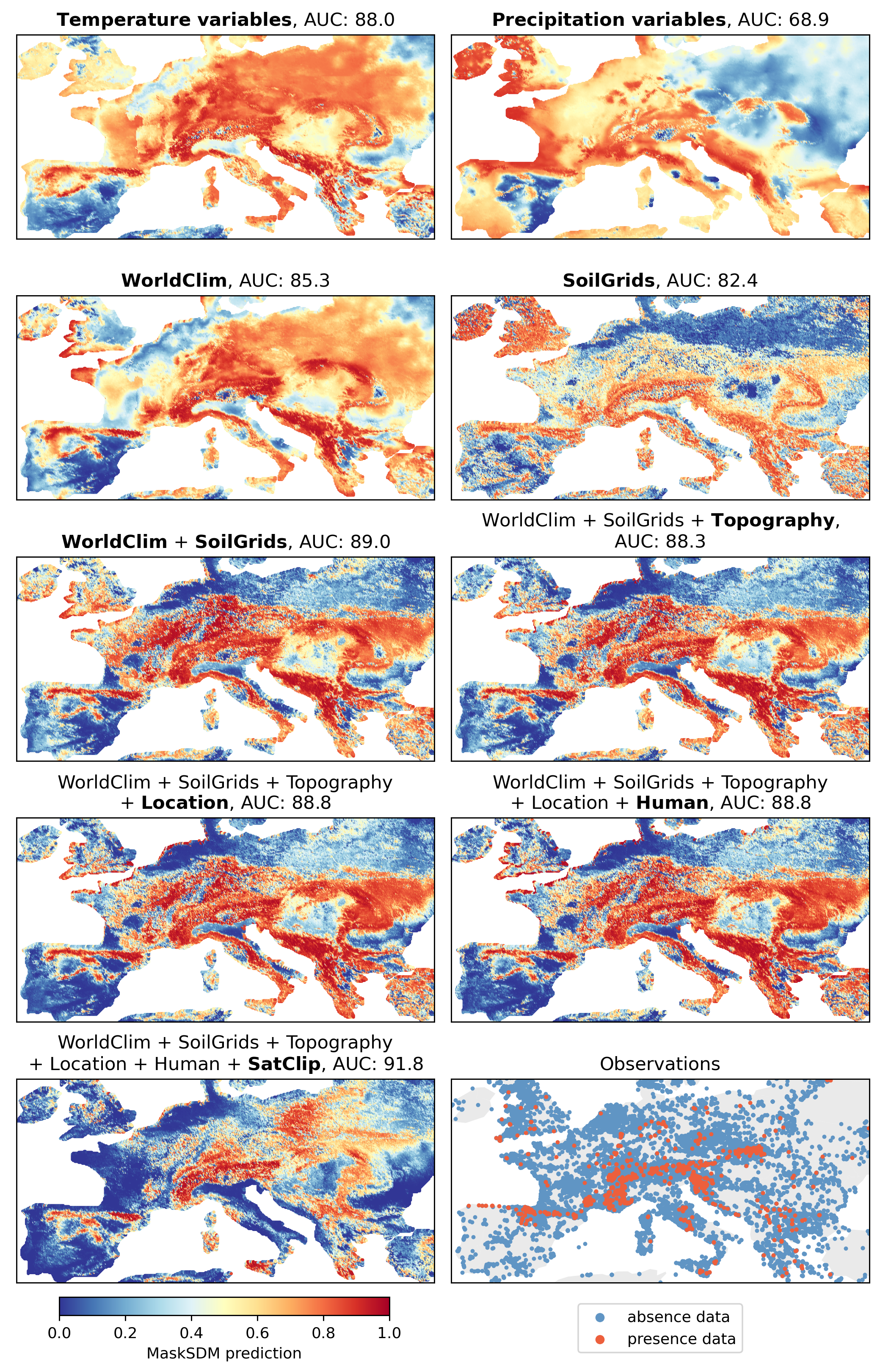}
    \caption{MaskSDM predicted suitability maps for kidney vetch (\textit{Anthyllis vulneraria}) using different subsets of input variables. For each subset, we report the corresponding AUC obtained for \textit{A. vulneraria} in the test set. The bottom-right panel shows the geographic distribution of observations, with presence data marked in red and absence data in blue.}
    \label{fig:prediction_maps}
\end{figure}

We present prediction maps of \textit{A. vulneraria}, generated using nine different subsets of variables, to examine spatial variations (Fig. \ref{fig:prediction_maps}). Prediction maps for \textit{V. myrtillus} and \textit{Q. ilex} are provided in Appendix \ref{sec:appendix_prediction_maps}. Both temperature and precipitation variables alone produce relatively coarse and contrasting patterns. In particular, Eastern Europe appears to be suitable for \textit{A. vulneraria} based on temperature alone, whereas this area is excluded by precipitation-based suitability predictions. Conversely, Southern Spain is found unsuitable based on temperature and suitable based on precipitation. However, both broadly align with the actual geographic range of presence observations, particularly in the Alps. Interestingly, the prediction map with WorldClim variables is not a simple linear combination of temperature and precipitation predictions, highlighting the limitations of overly simplistic linear models. Comparing WorldClim to SoilGrids, we notice differences in pattern resolution, with SoilGrids exhibiting more localized variations. Additionally, soil properties suggest that the northern regions in the map are unsuitable for \textit{A. vulneraria}. When WorldClim and SoilGrids are combined, the AUC improves significantly, and the resulting map more closely resembles the final prediction with all variables. This suggests that these two groups of variables are key determinants of the distribution of \textit{A. vulneraria}, consistent with previous findings by \cite{daco2021altitude}. In contrast, adding topography, location, and human-related variables does not substantially alter the prediction maps. This results in a slight decrease in AUC performance, possibly due to overfitting or spurious correlations for this species. These findings suggest that these variables have little influence on the species' distribution and could potentially be excluded. Ultimately, incorporating satellite image features narrows the predicted suitable areas, increasing the AUC. By examining these maps, users of MaskSDM can determine which prediction maps are most relevant to their needs and study area, and, consequently, which variables should be considered in the end.

\subsection{Explaining MaskSDM performance with Shapley values}

While analyzing predictions and performance provides valuable insights into model behavior with subsets of variables, it remains an indirect measure of predictor importance. To simplify the analysis and better understand the model's reliance on correlated sources of information, it is essential to summarize the contribution of individual predictors or groups of predictors with one concise value that disentangles their effects. Following this intuition, we leverage MaskSDM with the stratified Monte Carlo approach to calculate Shapley values for each predictor. We show that this approach yields faster and more stable estimates than the uniform Monte Carlo approach (Figs \ref{fig:shapley_values_convergence_1} and \ref{fig:shapley_values_convergence_2} in Appendix \ref{sec:appendix_shapley}). 

\begin{figure}
    \centering
    \input{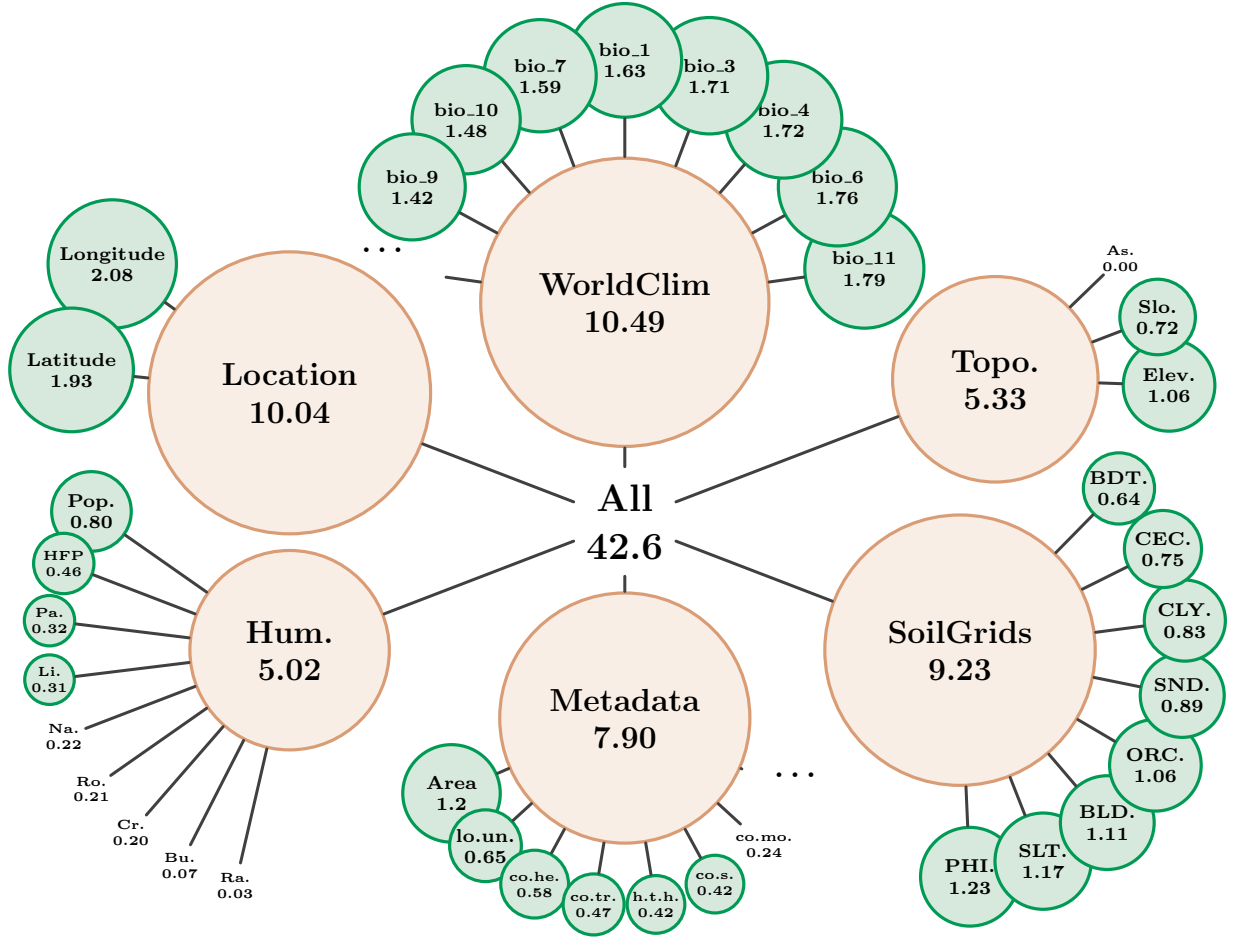}
    \caption{Shapley values explaining global AUC performance across all species on the test set, indicating the average contribution of individual predictors and predictor groups to the global performance. Shapley values for individual predictors (in green) can be compared against each other, while the Shapley value of predictors groups (in orange) can be compared among themselves.
    The size of each bubble is proportional to its corresponding Shapley value, with bubbles representing values below 0.25 omitted for clarity. Table \ref{tab:predictors} provides the complete list of Shapley values and predictor abbreviations.
    }
    \label{fig:shapley_tree}
\end{figure}

Analyzing the Shapley values for the six selected predictor groups, the WorldClim variables have the highest impact, contributing the most to the model performance on average (see Fig. \ref{fig:shapley_tree} and Table \ref{tab:predictors} for the exhaustive list of variables contributions). The location information and the SoilGrids predictors follow, which is expected given the high performance of the models based solely on location \citep{cole2023spatial}, and the importance of soil properties for plant ecology. Interestingly, the Shapley values for the group of metadata are relatively high, likely because variables such as plot size and vegetation cover provide unique, less correlated information compared to other groups. In contrast, human-related and topographic predictors have lower Shapley values, possibly because their influence is very localized.

Focusing on individual predictors, the highest Shapley values are achieved by longitude and latitude, suggesting that, despite their correlation with many other variables, precise location information provides significant additional predictive power.
WorldClim variables also have high and relatively similar Shapley values. Notably, temperature-related variables consistently rank higher than precipitation-related ones. In particular, the mean temperature of the coldest quarter (bio\_11) and the minimum temperature of the coldest month (bio\_6) stand out, showing that lower thermal limits play a crucial role in plant distribution. 
Surprisingly, the Shapley value of the aspect is zero, possibly because the fine-scale variations in aspect may be too noisy to be accurately interpreted. This highlights the utility of computing Shapley values, not only for assessing variable importance but also for identifying potential issues in the modeling pipeline.

\subsection{Explaining MaskSDM predictions with Shapley values}

\begin{figure}
    \centering
    \includegraphics[width=1\linewidth]{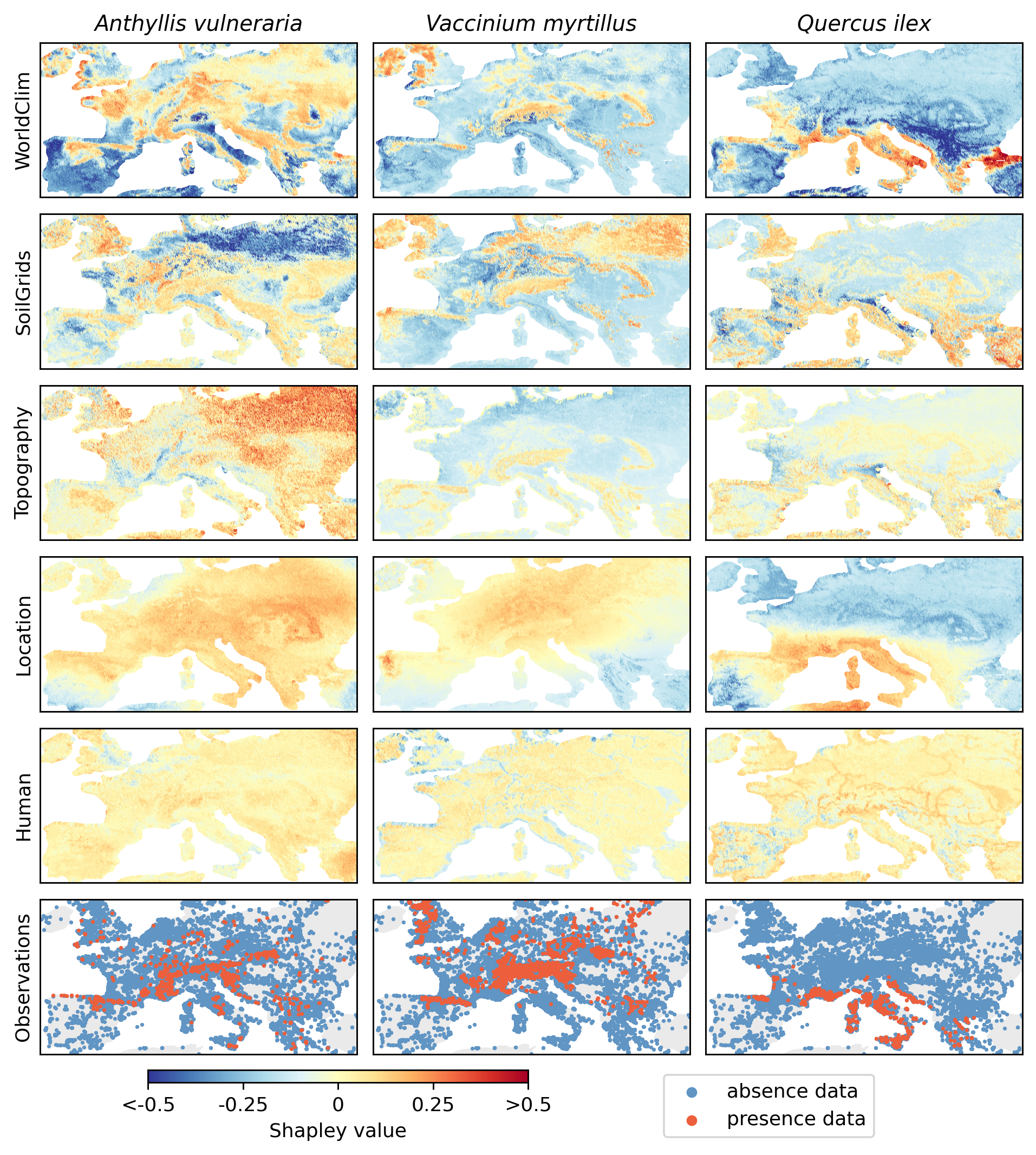}
    \caption{Shapley value maps representing the contribution of each predictor group to the MaskSDM predictions of kidney vetch (\textit{Anthyllis vulneraria}), European blueberry (\textit{Vaccinium myrtillus}), and holm oak (\textit{Quercus ilex}). The geographic distribution of observations (presence data in red and absence data in blue) is represented. For each location, the sum of the Shapley values across all the predictor groups equals the model prediction. Higher Shapley values indicate that the corresponding predictor group generally increases the predicted suitability for the given location.}
    \label{fig:shapley_maps}
\end{figure}

We compute and map Shapley values spatially by applying Equation \ref{eq:shapley} at each location. This provides a local contribution of each predictor group to the prediction per grid cell, allowing us to visualize how these contributions vary in space (Fig. \ref{fig:shapley_maps}). Comparing the Shapley value maps (Fig. \ref{fig:shapley_maps}) to the prediction maps (leftmost column of Fig. \ref{fig:prediction_maps}) for \textit{A. vulneraria} across multiple predictor groups, we gain spatially explicit insights into the relative importance of predictors with the predicted species suitability. As expected, WorldClim and SoilGrids exhibit similar overall patterns but with some localized specificities. For instance, along the west coast of Greece, the Shapley values for WorldClim differ significantly from the corresponding predictions, which indicate a highly suitable area. This discrepancy may arise from correlations among multiple predictors that drive high predicted suitability, even if climate may not be the primary factor in making the region particularly suitable for the species. 

For \textit{V. myrtillus} (center column of Fig. \ref{fig:shapley_maps}), although the distribution of observations is similar to that of \textit{A. vulneraria}, the Shapley values show important differences. In particular, soil properties appear to be more favorable for \textit{V. myrtillus} in northern regions and the proximity to water bodies (such as coastlines or rivers, as captured in the human influence variables) seems to constrain its range. This pattern contrasts with \textit{Q. ilex} (rightmost column of Fig. \ref{fig:shapley_maps}), for which proximity to water appears to increase suitability. Additionally, \textit{Q. ilex} distribution is significantly constrained by bioclimatic variables, underscoring their key role in shaping its range \citep{de2016quercus}. All these findings illustrate the inter-species differences in their response to various predictors, demonstrating how Shapley values can help quantify and explain these effects.

\section{Discussion}

Species distribution models have demonstrated their value in various ecological and conservation applications \citep{guisan2013predicting}. However, these diverse applications require flexibility in selecting input variables and the ability to analyze and explain model predictions. Additionally, SDMs often face challenges related to inconsistent environmental data between locations, making it difficult to transfer models to new areas with limited data availability \citep{petitpierre2017selecting}. In this work, we introduce MaskSDM, a deep learning approach based on masked data modeling that overcomes these limitations by allowing the model to consider an arbitrary subset of predictors as input while providing accurate explainability metrics on their contributions. We show that its predictions closely match those of a model trained specifically on the chosen subset of predictors and are more accurate than alternative approaches commonly used in SDMs. Additionally, MaskSDM allows the input variable selection to be tailored to specific locations, applications, and species, while maintaining the simplicity and effectiveness of a single model. MaskSDM also improves the interpretability of SDMs by enabling a clearer understanding of how different predictors contribute. 
Furthermore, it facilitates more reliable Shapley value estimation, providing a single score for each predictor or predictor group to summarize its contribution effectively. 
Unlike prediction maps, which may primarily reflect the spatial autocorrelation of species distributions, the computation of spatially explicit Shapley values highlights the influence of specific predictors, potentially offering a clearer view of the underlying biological processes \citep{fourcade2018paintings}. However, Shapley values have known limitations in establishing true causal relationships \citep{kumar2020problems} and should therefore be interpreted as insights into model behavior and ecological patterns.

While we test MaskSDM on a presence-absence dataset, the approach is not limited to a specific data type and can accommodate various types of species data. In particular, MaskSDM could easily be used with the growing volume of presence-only data from citizen science platforms, enabling the applicability of the model to a wider range of species with a larger number of observations \citep{cole2023spatial}. Additionally, our study focuses on tabular data and vector representations from satellite images as input. However, deep learning facilitates the integration of diverse data types, including time series, images, and textual information \citep{mizrahi20244m}. Since our approach is inherently multimodal, these different modalities can be encoded through tokenization and incorporated into the model. As a future work, we plan to expand MaskSDM to include these additional data sources, further improving the accuracy of species distributions. However, adding more predictors is not always beneficial and can sometimes lead to overfitting, reducing generalization capability, especially for species with fewer observations. We observe this with \textit{Anthyllis vulneraria}, where test set performance decreases when topography variables are added. 
Importantly, MaskSDM can easily be evaluated on any subset of variables without retraining, making the process more computationally efficient. This is particularly advantageous for iterative procedures such as stepwise variable selection \citep{williams2012environmental}, which can be used at inference to identify the optimal set of predictors (see Appendix \ref{appendix:workflow}). However, it remains essential to complement data-driven selection with expert knowledge to ensure ecological relevance. Beyond predictor selection, MaskSDM’s modular design also opens the door to further extensions. For instance, it could be extended with methods for uncertainty estimation—an essential aspect when working with ecological data—or with spatially explicit approaches that capture dependencies between sites \citep{ovaskainen2016uncovering, chen2019uncertainty, domisch2019spatially}. Furthermore, although biotic interactions were not considered in this study, they could be included either directly as predictors or through more advanced techniques such as graph neural networks \citep{strydom2023graph}.

An important concept in machine learning is \textit{pre-training}, which involves first training a model on a large, diverse dataset to learn generalizable representations before being adapted to specific tasks. This is useful when label data for a given task is scarce, as the model can leverage knowledge acquired from a broader dataset. This concept has been embodied by \textit{foundation models} \citep{bommasani2021opportunities}, which are trained on massive datasets—requiring tremendous computational resources—to capture broad, transferable patterns across multiple domains. Once such models are pre-trained, they can be fine-tuned on task-specific datasets, where additional labeled data helps the model specialize while retaining its generalization ability. This paradigm could be cautiously applied to SDMs: MaskSDM could serve as a pre-trained model, which could then be fine-tuned for specific regions or species of interest. For instance, a researcher might have additional observations for a particular species and could fine-tune MaskSDM to leverage both its broad generalization ability and the added specificity from new data. This idea aligns with the \textit{N-SDM} framework \citep{guisan2025spatially}, which integrates global and regional SDMs through a spatially-nested approach that considers scale-specific species and predictor data. However, instead of maintaining separate global and regional models, MaskSDM allows for a more seamless transition: it can be pre-trained on a global dataset—potentially including locations with fewer predictors—while incorporating data from data-rich regions either during pre-training or fine-tuning. Moreover, if additional predictors become available for a region at inference time, MaskSDM can easily integrate them through fine-tuning, by simply adding a new tokenizer, without having to re-train the whole system (see the workflow of Appendix \ref{appendix:workflow}). These promising directions suggest that MaskSDM could evolve into a foundation model for SDMs, offering a general pipeline that adapts to diverse ecological and environmental modeling needs.

\section*{Acknowledgements}

This work was supported by the Swiss National Science Foundation, under grant 200021\_204057 ``Learning unbiased habitat suitability at scale with AI (deepHSM)''.

\section*{Author contributions}
We adhere to the CRediT taxonomy.
\textbf{Robin Zbinden:} Conceptualization; Data Curation; Formal Analysis; Investigation; Methodology; Software; Validation; Visualization; Writing – Original Draft Preparation; Writing – Review \& Editing. 
\textbf{Nina van Tiel:} Conceptualization; Writing – Review \& Editing.
\textbf{Gencer Sumbul:} Conceptualization; Methodology; Supervision; Writing – Review \& Editing.
\textbf{Chiara Vanalli:} Conceptualization; Supervision; Writing – Review \& Editing.
\textbf{Benjamin Kellenberger:} Conceptualization; Funding Acquisition; Supervision.
\textbf{Devis Tuia:} Conceptualization; Funding Acquisition; Project Administration; Supervision; Writing – Review \& Editing.




\selectlanguage{english}
\FloatBarrier
\sloppy
\phantomsection

\newpage

\bibliography{references}
\bibliographystyle{abbrvnat}

\newpage

\appendix


\section{Computing Shapley values for a larger number of predictors}
\label{sec:appendix_shapley}

As explained in Section \ref{sec:shapley}, computing Shapley values requires evaluating the model an exponential number of times with respect to the number of predictors, i.e., $O(2^M)$, since it must consider all possible subsets of predictors. When dealing with only a few predictors or groups of predictors, this computation remains tractable. However, as the number of predictors increases, the computational cost quickly becomes prohibitive. To address this, Shapley values are typically estimated using Monte Carlo (MC) methods, which approximate their values by sampling and evaluating only $k$ subsets of predictors instead of $2^M$. These estimates converge as the number of sampled subsets increases.

\begin{wraptable}{r}{0.18\textwidth} 
    \centering
    \renewcommand{\arraystretch}{1.2}
    \setlength\tabcolsep{5pt}
    \small 
    \begin{tabular}{|c|c|c|c|c|}
        \hline
        3 & 1 & 5 & 4 & 2 \\ \hline
        4 & 2 & 1 & 5 & 3 \\ \hline
        5 & 4 & 3 & 2 & 1 \\ \hline
        2 & 5 & 4 & 1 & 3 \\ \hline
        1 & 3 & 2 & 4 & 5 \\ \hline
    \end{tabular}
    \caption{Example of a randomly generated $5 \times 5$ Latin square. Each number appears exactly once per row and column.}
    \label{tab:latin_square}
\end{wraptable}

\begin{figure}[tb]
    \centering
    \includegraphics[width=1\linewidth]{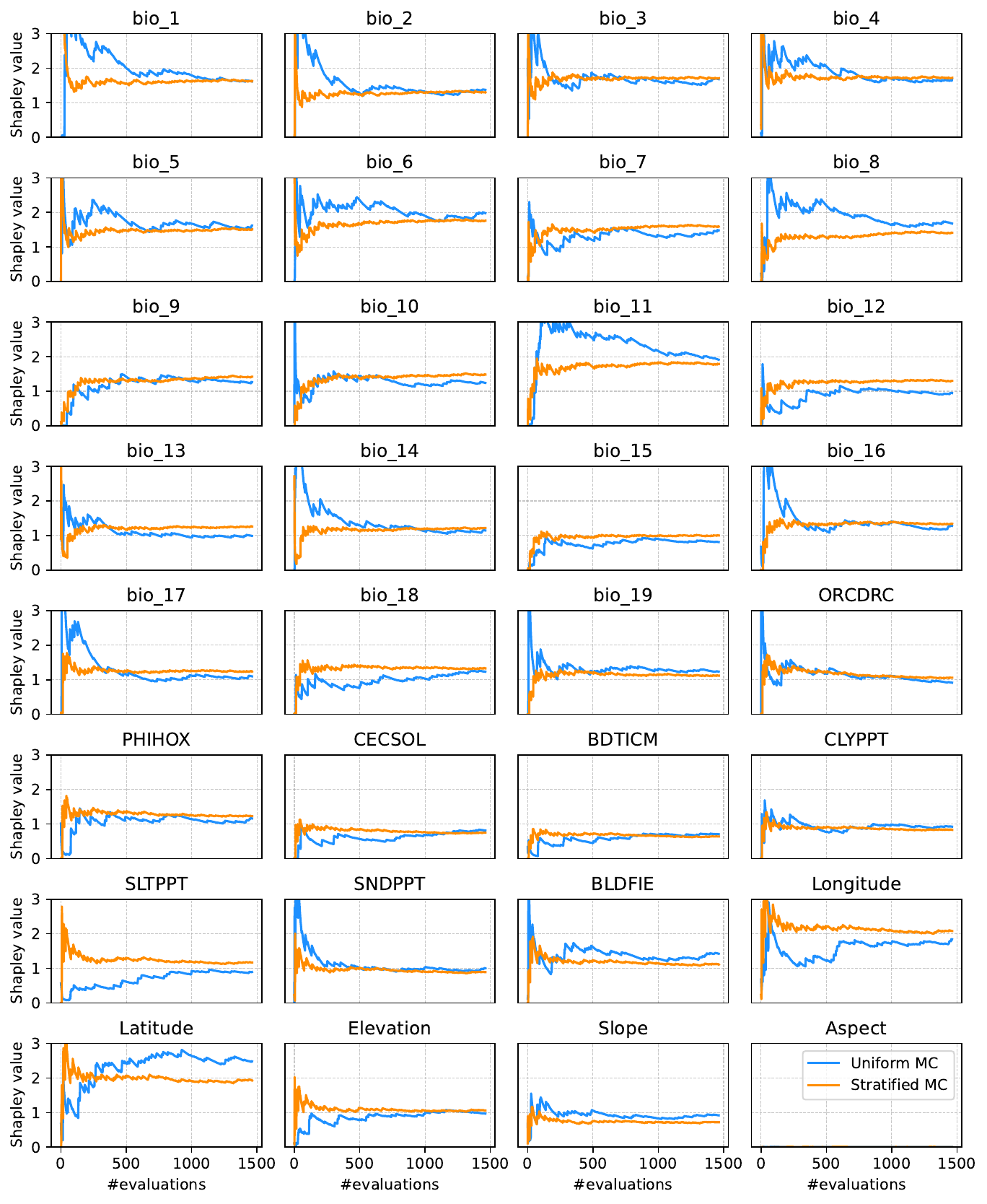}
    \caption{Shapley value convergence of the uniform and stratified Monte Carlo approaches (WorldClim, SoilGrids, and topography predictors). Note: apparently empty plots indicate Shapley values close to zero.}
    \label{fig:shapley_values_convergence_1}
\end{figure}

\begin{figure}[tb]
    \centering
    \includegraphics[width=1\linewidth]{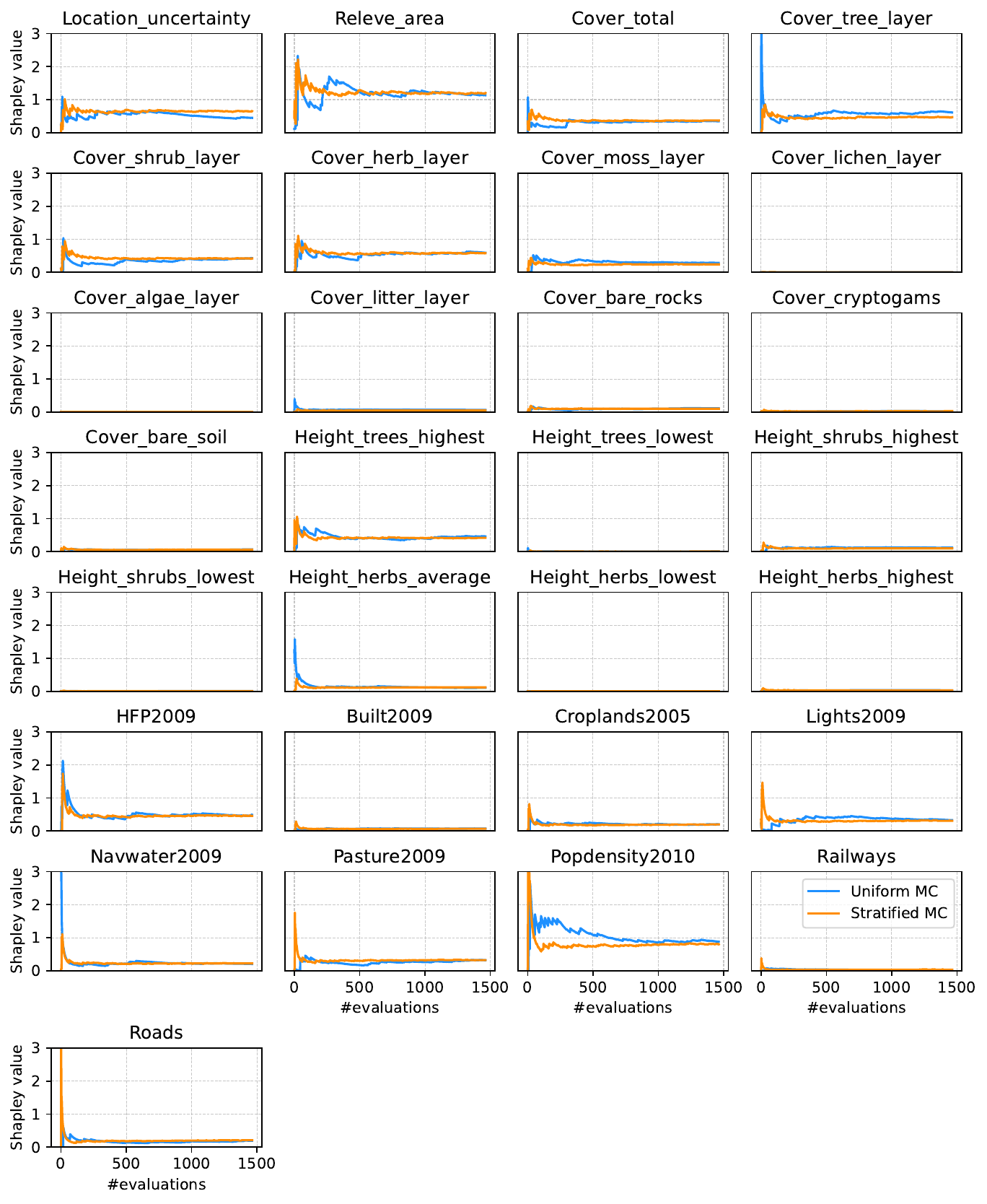}
    \caption{Shapley value convergence of the uniform and stratified Monte Carlo approaches (metadata and human influence predictors). Note: apparently empty plots indicate Shapley values close to zero.}
    \label{fig:shapley_values_convergence_2}
\end{figure}

However, in practice, uniform random sampling of predictor subsets often leads to poor estimates, particularly in our application. A key issue is that adding a single predictor to the empty set results in a significant AUC improvement, often increasing from 50\% to as much as 80\%. In contrast, adding additional predictors yields only marginal performance gains. Consequently, the estimates are highly sensitive to how frequently the empty set is sampled, leading to high variance and slow convergence. To mitigate this issue, we adopt a stratified MC approach, in contrast to uniform MC sampling. In this approach, each subset size is sampled an equal number of times, ensuring that the empty set is considered at the same frequency as other subset sizes. Since each term in Equation \ref{eq:shapley} requires two model evaluations, we also develop an optimization strategy to reuse model predictions, significantly reducing computational cost. Inspired by \citep{covert2020understanding}, our method sequentially adds one variable at a time while preserving the stratified sampling structure.
To implement this, we leverage \textit{Latin squares} \citep{keedwell2015latin}, which are square matrices where each element appears exactly once per row and column (example in Table \ref{tab:latin_square}). Our approach, outlined in Algorithm \ref{algo:shapley}, assigns predictor indices (ranging from $\{1,\ldots,M\}$) as elements of the Latin square. By reading each row of the matrix, we define the order in which predictors are added, ensuring that every predictor appears in every possible position. The Latin square is randomly sampled, so each row follows a different variable ordering.
After completing a full Latin square, each predictor has been considered in every position, meaning that  $M$  terms have been computed for each Shapley value. This process can be repeated $N$ times by generating different Latin squares. We compare the convergence rates of uniform and stratified methods in Figs \ref{fig:shapley_values_convergence_1} and \ref{fig:shapley_values_convergence_2}, showing that the stratified approach achieves more stable and faster convergence.
\begin{algorithm}
\caption{Stratified Monte Carlo Shapley Value Estimation using Latin Squares}
\label{algo:shapley}
\begin{algorithmic}[1]
    \Require Predictor set size $M$, number of Latin squares $N$, performance metric $f$ of MaskSDM model
    \Ensure Estimated Shapley values $\hat{\phi}_1, \hat{\phi}_2, \ldots, \hat{\phi}_{M}$
    \State Initialize $\hat{\phi}_1 \gets 0, \hat{\phi}_2 \gets 0, \ldots, \hat{\phi}_{M} \gets 0$
    \For{$n = 1$ to $N$} 
        \State Generate a random $M \times M$ Latin square $L$
        \For{each row $r$ in $L$} \Comment{Iterate over Latin square rows}
            \State $S \gets \emptyset$ \Comment{Initialize subset}
            \State $f_{\text{prev}} \gets f(S)$ \Comment{Compute performance for empty set ($\approx 0.5$ for AUC)}
            \For{$j = 1$ to $M$} 
                \State $x_i \gets L[r, j]$ \Comment{Select predictor from Latin square}
                \State $S \gets S \cup \{x_i\}$ \Comment{Add predictor to subset}
                \State $f_{\text{curr}} \gets f(S)$ \Comment{Evaluate model performance with updated subset}
                \State $\hat{\phi}_i \gets \hat{\phi}_i + (f_{\text{curr}} - f_{\text{prev}})$ \Comment{Update Shapley estimate}
                \State $f_{\text{prev}} \gets f_{\text{curr}}$ \Comment{Store current performance for next iteration}
            \EndFor
        \EndFor
    \EndFor
    \State \Return $\frac{\hat{\phi}_1}{NM}, \frac{\hat{\phi}_1}{NM}, \ldots, \frac{\hat{\phi}_M}{NM}$
\end{algorithmic}
\end{algorithm}

\section{MaskSDM Workflow}
\label{appendix:workflow}

In this section, we outline how we envision MaskSDM being applied, highlighting the different scenarios in which it could be used. We emphasize that MaskSDM is a flexible method rather than a single fixed model, and it can be adapted to various contexts. For this reason, we propose a workflow to guide its use. The workflow consists essentially of two stages: a training stage and an inference stage, where the model can potentially be fine-tuned depending on the specific scenario. Consequently, the inference stage may differ according to the application context. Additionally, practitioners may engage with the workflow at different levels of expertise. Typically, the training stage could be handled by those with coding and modeling experience, while the trained model could then be made accessible to stakeholders who may have less technical expertise but bring valuable ecological knowledge. In the inference stage, these stakeholders can interact with the model by providing different sets of predictors and examining how predictions, performance, and predictor importance change accordingly.

\subsection{Training Stage}

The training stage aims to build a robust MaskSDM model capable of handling any subset of predictors using the masking procedure described in this paper. While we illustrate this process using the sPlotOpen dataset with a specific set of predictors, the approach can be applied to any sufficiently large dataset. At this stage, we recommend using the largest possible pool of species and predictors to maximize the model’s flexibility during inference. However, care may be required to address data imbalances arising from different types of sampling biases \citep{zbinden2024selection}. The ultimate goal is to develop a model that is as generalizable as possible, capable of supporting diverse species, spatial extents, and applications.

\subsection{Inference stage}

MaskSDM can be deployed in multiple contexts. The simplest case is to use the model under the same setup as during training, i.e., the same species, region, and set of predictors. In this situation, MaskSDM yields strong predictive performance and allows for accurate estimation of predictor importance. In practice, however, users may wish to focus on particular species or incorporate additional ecological knowledge about which predictors are most relevant. In such cases, predictors considered irrelevant can be excluded from the inputs. Alternatively, if the goal is to reduce the dimensionality of the predictor set, users may apply a variable selection method and retain only a subset of predictors to compute the final predictions. MaskSDM naturally supports this and provides predictions equivalent to a model trained directly with fewer predictors (see Table \ref{tab:baselines}). Shapley values can also be leveraged to better identify which predictors are most relevant to retain. Another possibility is to train a MaskSDM model in a data-rich region and then transfer it to a region with fewer available data. If the inference region has fewer predictors available, MaskSDM can seamlessly adapt by operating only on the provided subset. Other situations may arise where adapting the model to new data is required. We discuss such cases below.

\subsubsection{Additional species observations}

An end-user may possess a small number of additional species observations—possibly private data—that they wish to leverage to improve predictions. In this case, the model can be fine-tuned on these observations, i.e., trained for a few additional steps to better fit the new data. This type of adaptation is commonly referred to as \textit{few-shot learning} \citep{lange2025feedforward}. There are multiple strategies for doing so. One option is to fine-tune all model weights, though this may require more data. A lighter alternative is to reuse the environmental representation learned in the last hidden layer (which already captures general ecological characteristics) and only update the weights of the final prediction layer. More advanced approaches may yield further improvements, but these two options are straightforward and effective.

\subsubsection{Additional predictors}

End-users may also want to incorporate predictors that were not available during the original training but are considered important. In such cases, the MaskSDM framework can be extended by adding a tokenizer for the new predictor that produces a corresponding token representation. However, this addition requires retraining or fine-tuning the model with species observations, since the model parameters must learn how to integrate the new predictor into the existing representation.

\subsubsection{Additional species}

Finally, one may wish to add entirely new species that were not part of the training pool. Naturally, this requires the availability of occurrence records for the new species. The situation is similar to the case of additional species observations: one can reuse the general environmental representation extracted by the last shared layer and train a lightweight prediction head (e.g., a linear layer) to model the distribution of the new species. This strategy leverages the fact that strong environmental features have already been extracted and are broadly useful across species. If the new species is ecologically similar to those in the training set, this approach should work well. However, if the species differs substantially, it may be necessary to also incorporate additional predictors, as discussed in the previous subsection.

\section{Additional details on experimental setup}

\subsection{Model architecture and training details}
\label{sec:architecture_details}

The tabular predictors are tokenized using periodic activation functions \citep{sitzmann2020implicit, gorishniy2022embeddings} defined as:
\begin{align}
    f_i(x) = \texttt{concat}[\text{sin}(v), \text{cos}(v)], \quad v= [2\pi c_1 x_i, \ldots, 2\pi c_k x_i]
\end{align}
where $x_i$ represents the scalar value of the $i$-th tabular variable, $k=48$ is the number of frequencies \citep{gorishniy2022embeddings}, and $c_i$ are trainable parameters. The output of these periodic activation functions is passed through a linear layer followed by a ReLU activation.  This encoding method has been shown to improve numerical data representation \citep{gorishniy2022embeddings} and is particularly effective at capturing multi-scale patterns in geographic coordinates \citep{russwurm2024locationencoding}.
Satellite image features are obtained using the SatCLIP encoder \citep{klemmer2023satclip}, which employs \num{40} Lagrange polynomials and is distilled from the ViT-B/16 model \citep{dosovitskiy2020image} trained on Sentinel-2 imagery. This encoder (with frozen weights) produces an embedding that is projected into the space of tokens using a linear layer. The resulting token is added to the tokens obtained from tabular data variables.

The transformer encoder is based on the architecture of the FTTransformer \citep{gorishniy2021revisiting} and consists of \num{7} identical blocks. Each block processes tokens of size \num{192}, with the number of tokens equal to the number of input variables, that is, \num{62}. These blocks incorporate self-attention with \num{8} heads and a feed-forward network, interleaved with layer normalization and dropout (with a probability of \num{0.1}). The outputs of the final transformer block are aggregated using average pooling to produce a single vector of size \num{192}. A linear prediction head with a sigmoid activation function generates suitability scores for the \num{12738} species.

The model is trained using the schedule-free AdamW optimizer \citep{loshchilov2017decoupled, defazio2024road} with the following hyperparameters: a learning rate of \num{0.001}, \num{1000} warm-up steps, weight decay of \num{0.01}, and a batch size of \num{256}. To address the imbalance between presence and absence data (see section \ref{sec:appendix_imbalance}), we employ a weighted binary cross-entropy loss for multi-label classification. Species-specific weights are computed following the method proposed by \cite{zbinden2024selection}. Training is performed for a maximum of \num{1000} epochs, with early stopping\footnote{the model parameters are saved at the point of highest performance} based on the area under the receiver operating characteristic curve (AUC) on the validation set.

\subsection{Baseline implementation details}
\label{sec:baseline_details}

We provide additional details on the baselines compared to MaskSDM. The first two approaches replace missing values with the arithmetic \textit{mean} or \textit{median} of the corresponding predictor. These methods minimize the squared or absolute difference, respectively, on average.  However, they can significantly bias predictions, particularly when the true values deviate substantially from these global statistics. More sophisticated imputation methods leverage the training distribution to estimate missing values. Our third baseline samples missing values from the \textit{marginal} distribution, i.e., the empirical distribution of training samples \citep{lundberg2017unified}. This involves randomly selecting training samples for each inference instance, substituting missing values with the corresponding values from the sampled training data, repeating this process $m$ times, and averaging the resulting predictions. However, this method assumes predictor independence, which rarely holds in SDMs. The final imputation approach uses the \textit{conditional} distribution. This involves finding the $m$ nearest neighbors of the inference sample within the training set, a computationally expensive operation. Missing values are then replaced by those from the nearest neighbors, and the $m$ predictions are averaged, similar to the marginal distribution imputing baseline. In our experiments, we set $m=100$ for the marginal distribution and $m=5$ for the conditional distribution. Both require access to the training set during inference and increase inference time by a factor of $m$. 

For all five baselines (oracle, mean, median, marginal, and conditional imputing), we employ the same architecture and training procedure as MaskSDM but without masking. During training, the missing variables are replaced by the mean (or median for the median imputing baseline) of the corresponding variable.

\subsection{Geographic distribution of plots}
\label{sec:appendix_dataset}

\begin{figure}[H]
    \centering
    \includegraphics[width=1\linewidth]{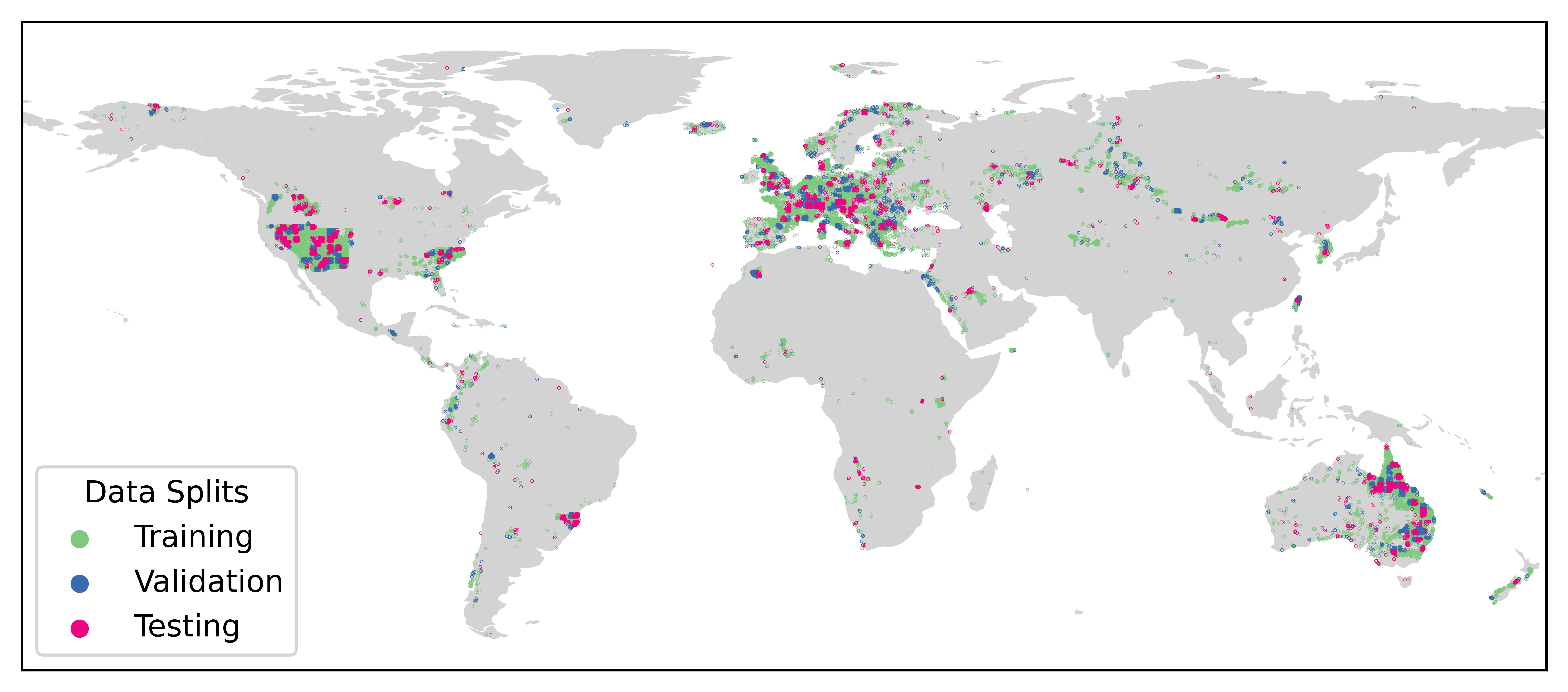}
    \caption{Geographic distribution of sPlotOpen plots across training, validation, and testing splits generated via spatial block cross-validation.}
    \label{fig:splits}
\end{figure}

Fig. \ref{fig:splits} illustrates the geographic distribution of sPlotOpen (presence-absence) plots used for training, hyperparameter tuning, and testing in our MaskSDM model and baselines comparison. The dataset is split using spatial block cross-validation \citep{roberts2017cross}, ensuring that the training, validation, and testing sets do not overlap geographically.  

\subsection{Distribution of presence records}
\label{sec:appendix_imbalance}

\begin{figure}[H]
    \centering
    \includegraphics[width=1\linewidth]{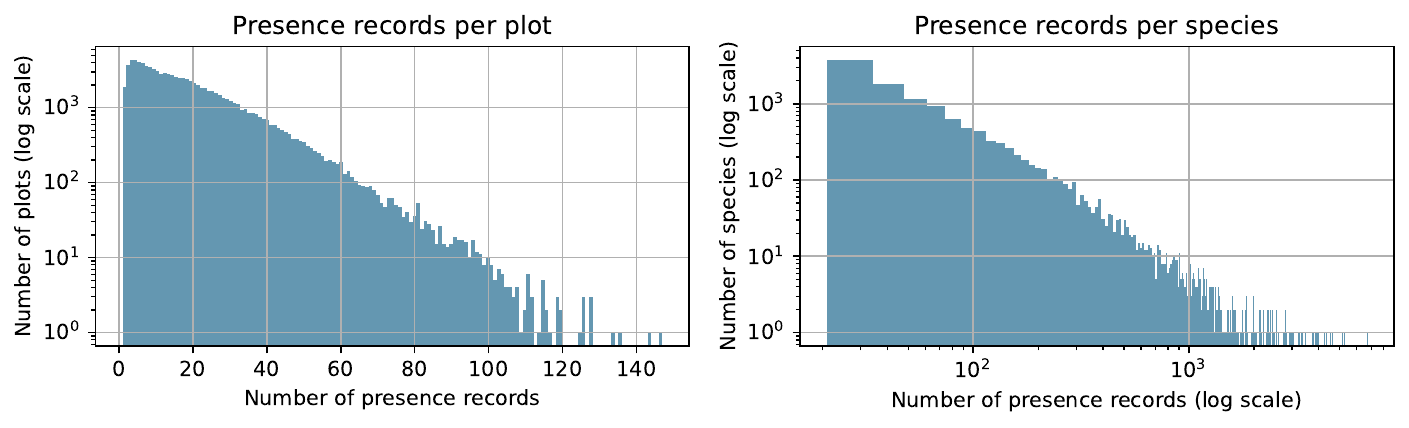}
    \caption{Distribution of presence records per plot and per species used in the experiments.}
    \label{fig:presence_records}
\end{figure}

The total number of presence records (across training, validation, and test sets) follows a long-tailed distribution across both plots and species, as shown in Fig. \ref{fig:presence_records}.

\section{Additional results}
\label{sec:appendix_results}

\subsection{Performance across training epoch}
\label{sec:appendix_epoch}

Table \ref{tab:epoch} reports the mean test AUC of MaskSDM across different training epochs. After only five epochs, MaskSDM already outperforms the mean imputation baseline in most cases, except when all predictors are used. By epoch 25, it achieves comparable performance to the baseline even in the all-predictor setting. We also find that the performance improvements from additional training epochs are more pronounced when fewer predictors are available. Overall, MaskSDM shows no evidence of overfitting, likely due to the regularizing effect of the masking procedure.

\begin{table}
    \renewcommand{\arraystretch}{\stretchtable}
    \setlength{\arrayrulewidth}{0.2pt}
    \setlength\tabcolsep{5pt}
    \centering
    \small
    \begin{tabular}
        {|l|l|cccccccccc|}
        \hline
        \multirow{8}{*}{\rotatebox[origin=c]{90}{\textbf{Predictors (\#)}}} & 
        Avg. Temperature (1) & \cmark & \xmark & \xmark & \cmark & \cmark & \cmark & \cmark & \cmark & \cmark & \cmark \\
        & \cellcolor{mygray}WorldClim (19) & \cellcolor{mygray}\xmark & \cellcolor{mygray}\xmark & \cellcolor{mygray}\xmark & \cellcolor{mygray}\cmark & \cellcolor{mygray}\cmark & \cellcolor{mygray}\cmark & \cellcolor{mygray}\cmark & \cellcolor{mygray}\cmark & \cellcolor{mygray}\cmark  & \cellcolor{mygray}\cmark \\
        & SoilGrids (8) & \xmark & \xmark & \xmark & \xmark & \cmark & \cmark & \cmark & \cmark & \cmark  & \cmark \\
        & \cellcolor{mygray}Topographic (3) & \cellcolor{mygray}\xmark & \cellcolor{mygray}\xmark & \cellcolor{mygray}\xmark & \cellcolor{mygray}\xmark & \cellcolor{mygray}\xmark & \cellcolor{mygray}\cmark & \cellcolor{mygray}\cmark & \cellcolor{mygray}\cmark &  \cellcolor{mygray}\cmark  & \cellcolor{mygray}\cmark\\
        & Location (2) & \xmark & \cmark & \xmark & \xmark & \xmark & \xmark & \cmark & \cmark & \cmark  & \cmark\\
        & \cellcolor{mygray}Human footprint (9)  & \cellcolor{mygray}\xmark & \cellcolor{mygray}\xmark & \cellcolor{mygray}\xmark & \cellcolor{mygray} \xmark & \cellcolor{mygray}\xmark & \cellcolor{mygray}\xmark & \cellcolor{mygray}\xmark & \cellcolor{mygray}\cmark & \cellcolor{mygray}\cmark  & \cellcolor{mygray}\cmark \\
        & Plot metadata (20) & \xmark & \xmark & \xmark & \xmark & \xmark & \xmark & \xmark & \xmark & \cmark  & \cmark \\
        & \cellcolor{mygray}Satellite image features & \cellcolor{mygray}\xmark & \cellcolor{mygray}\xmark & \cellcolor{mygray}\cmark & \cellcolor{mygray}\xmark & \cellcolor{mygray}\xmark & \cellcolor{mygray}\xmark & \cellcolor{mygray}\xmark & \cellcolor{mygray}\xmark & \cellcolor{mygray}\xmark  & \cellcolor{mygray}\cmark \\
         \hline
        \multirow{8}{*}{\rotatebox[origin=c]{90}{\textbf{MaskSDM epoch}}}
        & 5 & 82.0 & 90.9 & 87.5 & 95.0 & 95.5 & 95.5 & 96.1 & 96.0 & 96.2 & 96.3  \\
        & 10 & 84.2 & 93.7 & 92.2 & 96.3 & 96.6 & 96.7 & 97.0 & 97.0 & 97.1 & 97.3 \\
        & 25 & 84.8 & 95.0 & 94.2 & 96.7 & 97.1 & 97.1 & 97.3 & 97.3 & 97.6 & 97.8  \\
        & 50  & 85.4 & 95.4 & 95.2 & 96.9 & 97.2 & 97.3 & 97.5 & 97.5 & 97.8 & 97.9  \\
        & 100 & 85.1 & 95.8 & 95.7 & 97.0 & 97.3 & 97.4 & \textbf{97.6} & \textbf{97.6} & 97.9 & \textbf{98.0} \\
        & 178 & 85.1 & 96.0 & 95.7 & \textbf{97.1} & \textbf{97.4} & \textbf{97.5} & \textbf{97.6} & \textbf{97.6} & \textbf{98.0} & \textbf{98.0} \\
        & 500 & 85.4 & 96.2 & 95.9 & \textbf{97.1} & \textbf{97.4} & \textbf{97.5} & \textbf{97.6} & \textbf{97.6} & \textbf{98.0} & \textbf{98.0} \\
        & 1000 & \textbf{85.5} & 96.4 & \textbf{96.0} & \textbf{97.1} & \textbf{97.4} & \textbf{97.5} & \textbf{97.6} & \textbf{97.6} & \textbf{98.0} & 97.9 \\
         \hline
        \multirow{4}{*}{\rotatebox[origin=c]{90}{\textbf{Baselines}}}
        
        & Mean Imputing              & 69.6 & 77.8 & 75.6 & 88.7 & 91.3 & 91.8 & 95.5 & 95.5 & 95.9 & 97.8 \\
        & Median Imputing           & 73.5 & 82.4 & 74.7 & 89.4 & 91.1 & 91.6 & 95.9 & 95.8 & 96.3 & 97.9 \\
        & Marginal Imputing         & 62.4 & 79.8 & 82.1 & 89.5 & 92.7 & 93.2 & 96.1 & 96.1 & 96.6 & 97.8 \\ 
        & Conditional Imputing       & 72.6 & \textbf{96.6} & 95.0 & 96.6 & 97.0 & 97.0 & 97.2 & 97.0 & 97.6 & 97.8   \\ \hline
    \end{tabular}%
    \caption{Evolution of mean test AUC achieved by MaskSDM across epochs, compared to the imputing baselines. The highest validation AUC is reached at epoch \num{178}, and the corresponding model is used for the experiments. For the imputing baselines, results are reported at the epoch that maximizes validation AUC: 19 for mean, marginal, and conditional imputation, and 14 for median imputation. Bold values indicate the best performance in each column.}
    \label{tab:epoch}
\end{table}

\subsection{Performance by number of species}
\label{sec:appendix_occ}

Table \ref{tab:num_occ} presents the mean test AUC achieved by MaskSDM for groups of species categorized by their number of presence records (occurrences). We observe that the number of occurrences has a strong influence on performance. Moreover, the inclusion of additional predictors tends to provide greater benefits for species with more presence records.

\begin{table}
    \renewcommand{\arraystretch}{\stretchtable}
    \setlength{\arrayrulewidth}{0.2pt}
    \setlength\tabcolsep{5pt}
    \centering
    \small
    \begin{tabular}
        {|l|lc|cccccccccc|}
        \hline
        \multirow{8}{*}{\rotatebox[origin=c]{90}{\textbf{Predictors (\#)}}}
        & Avg. Temperature (1) & & \cmark & \xmark & \xmark & \cmark & \cmark & \cmark & \cmark & \cmark & \cmark & \cmark \\
        & \cellcolor{mygray}WorldClim (19) & \cellcolor{mygray} & \cellcolor{mygray}\xmark & \cellcolor{mygray}\xmark & \cellcolor{mygray}\xmark & \cellcolor{mygray}\cmark & \cellcolor{mygray}\cmark & \cellcolor{mygray}\cmark & \cellcolor{mygray}\cmark & \cellcolor{mygray}\cmark & \cellcolor{mygray}\cmark  & \cellcolor{mygray}\cmark \\
        & SoilGrids (8) & & \xmark & \xmark & \xmark & \xmark & \cmark & \cmark & \cmark & \cmark & \cmark  & \cmark \\
        & \cellcolor{mygray}Topographic (3) & \cellcolor{mygray} & \cellcolor{mygray}\xmark & \cellcolor{mygray}\xmark & \cellcolor{mygray}\xmark & \cellcolor{mygray}\xmark & \cellcolor{mygray}\xmark & \cellcolor{mygray}\cmark & \cellcolor{mygray}\cmark & \cellcolor{mygray}\cmark &  \cellcolor{mygray}\cmark  & \cellcolor{mygray}\cmark\\
        & Location (2) & & \xmark & \cmark & \xmark & \xmark & \xmark & \xmark & \cmark & \cmark & \cmark  & \cmark\\
        & \cellcolor{mygray}Human footprint (9) & \cellcolor{mygray} & \cellcolor{mygray}\xmark & \cellcolor{mygray}\xmark & \cellcolor{mygray}\xmark & \cellcolor{mygray} \xmark & \cellcolor{mygray}\xmark & \cellcolor{mygray}\xmark & \cellcolor{mygray}\xmark & \cellcolor{mygray}\cmark & \cellcolor{mygray}\cmark  & \cellcolor{mygray}\cmark \\
        & Plot metadata (20) &  & \xmark & \xmark & \xmark & \xmark & \xmark & \xmark & \xmark & \xmark & \cmark  & \cmark \\
        & \cellcolor{mygray}Satellite image features & \cellcolor{mygray} & \cellcolor{mygray}\xmark & \cellcolor{mygray}\xmark & \cellcolor{mygray}\cmark & \cellcolor{mygray}\xmark & \cellcolor{mygray}\xmark & \cellcolor{mygray}\xmark & \cellcolor{mygray}\xmark & \cellcolor{mygray}\xmark & \cellcolor{mygray}\xmark  & \cellcolor{mygray}\cmark \\
         \hline
        \multirow{6}{*}{\rotatebox[origin=c]{90}{\textbf{Species with}}}
        &    & \textbf{\#species}  &  &  &  &  &  &  &  &  &  &  \\
        & \#occ $>$ 20, i.e., all species      & 9009 & 85.1 & 96.0 & 95.7 & 97.1 & 97.4 & 97.5 & 97.6 & 97.6 & 98.0 & 98.0 \\
        & \#occ $>$ 1000    & 228  & 78.9 & 89.2 & 90.0 & 92.1 & 93.0 & 93.3 & 93.5 & 93.5 & 94.9 & 94.9 \\
        &  1000 $\geq$ \#occ $>$ 100 &  3377 & 84.9 & 95.3 & 95.6 & 96.9 & 97.3 & 97.4 & 97.5 & 97.5 & 97.9 & 97.9 \\ 
        &  100 $\geq$ \#occ $>$ 40 &  2956 & 85.6 & 96.5 & 96.3 & 97.5 & 97.7 & 97.8 & 97.9 & 97.9 & 98.2 & 98.2 \\ 
        & 40 $\geq$ \#occ $>$ 20 & 2448 & 85.6 & 96.7 & 95.5 & 97.4 & 97.6 & 97.7 & 97.8 & 97.8 & 98.2 & 98.2 \\\hline
    \end{tabular}%
    \caption{Mean test AUC comparison across species subsets grouped by the number of presence records (occurrences) in the training set.}
    \label{tab:num_occ}
\end{table}

\subsection{Performance with larger spatial blocks}
\label{sec:larger_blocks}

Table \ref{tab:larger_blocks} shows the mean test AUC of MaskSDM, the mean imputing baseline, and the oracle when the data are split using larger spatial blocks in block cross-validation, i.e., 5°$\times$5°, instead of 1°$\times$1°. Performance is computed on the subset of species with presences in all three splits, which differs from the species set used in other tables and is therefore not directly comparable. We find that MaskSDM again closely matches the oracle performance and, in some cases, even exceeds it, likely due to the regularizing effect of masking.

\begin{table}
    \renewcommand{\arraystretch}{\stretchtable}
    \setlength{\arrayrulewidth}{0.2pt}
    \setlength\tabcolsep{5pt}
    \centering
    \small
    \begin{tabular}
        {|l|l|cccccc|}
        \hline
        \multirow{7}{*}{\rotatebox[origin=c]{90}{\textbf{Predictors (\#)}}} 
        & Avg. Temperature (1) & \cmark & \xmark & \cmark & \cmark & \cmark  & \cmark \\
        & \cellcolor{mygray}WorldClim (19) & \cellcolor{mygray}\xmark & \cellcolor{mygray}\xmark & \cellcolor{mygray}\cmark & \cellcolor{mygray}\cmark & \cellcolor{mygray}\cmark & \cellcolor{mygray}\cmark \\
        & SoilGrids (8) & \xmark & \xmark & \xmark & \cmark & \cmark  & \cmark  \\
        & \cellcolor{mygray}Topographic (3) &  \cellcolor{mygray}\xmark & \cellcolor{mygray}\xmark & \cellcolor{mygray}\xmark & \cellcolor{mygray}\xmark & \cellcolor{mygray}\cmark  &\cellcolor{mygray}\cmark  \\
        & Location (2) &\xmark & \cmark & \xmark & \xmark & \cmark  &\cmark \\
        & \cellcolor{mygray}Human footprint (9) & \cellcolor{mygray}\xmark & \cellcolor{mygray}\xmark & \cellcolor{mygray}\xmark & \cellcolor{mygray}\xmark & \cellcolor{mygray}\xmark  & \cellcolor{mygray}\cmark  \\
        & Plot metadata (20) & \xmark & \xmark & \xmark & \xmark & \xmark  & \cmark \\
         \hline
         \multirow{6}{*}{\rotatebox[origin=c]{90}{\textbf{Method}}}
       & \textbf{Imputing}:        &  &  &  &  &  &  \\
        & Mean              & 66.4 & 67.3 & 86.8 & 89.6 & 93.7 & 96.0 \\ \cline{2-8}
        & \textbf{Masking}:        &  &  &   &  &  &  \\
        & MaskSDM (ours)    & \textbf{83.3} & \textbf{93.7} & \textbf{94.7} & \textbf{95.2} & \textbf{95.6} & \textbf{96.1} \\ \cline{2-8}
        & \textbf{Oracle}:        &  &  &  &  &  &  \\
        & One model per column & \textbf{83.3} & \textbf{93.7} & 94.3 & 95.1 & \textbf{95.6} & 96.0  \\ \hline
    \end{tabular}%
    \caption{Mean AUC performance of MaskSDM compared to the mean imputation baseline and the oracle when data are split using larger blocks in block cross-validation. Each block measures 5°$\times$5°, instead of the 1°$\times$1° blocks used elsewhere in this study. Mean performance is computed over the subset of species with presences in all three splits.}
    \label{tab:larger_blocks}
\end{table}

\subsection{Difference in predictions with Oracle}

\begin{table}
    \renewcommand{\arraystretch}{\stretchtable}
    \setlength{\arrayrulewidth}{0.2pt}
    \setlength\tabcolsep{4.8pt}
    \centering
    \small
    \begin{tabular}
        {|l|l|cccccccccc|}
        \hline
        \multirow{8}{*}{\rotatebox[origin=c]{90}{\textbf{Predictors (\#)}}} 
        & Avg. Temperature (1) & \cmark & \xmark & \xmark & \cmark & \cmark & \cmark & \cmark & \cmark & \cmark & \cmark \\
        & \cellcolor{mygray}WorldClim (19) & \cellcolor{mygray}\xmark & \cellcolor{mygray}\xmark & \cellcolor{mygray}\xmark & \cellcolor{mygray}\cmark & \cellcolor{mygray}\cmark & \cellcolor{mygray}\cmark & \cellcolor{mygray}\cmark & \cellcolor{mygray}\cmark & \cellcolor{mygray}\cmark  & \cellcolor{mygray}\cmark \\
        & SoilGrids (8) & \xmark & \xmark & \xmark & \xmark & \cmark & \cmark & \cmark & \cmark & \cmark  & \cmark \\
        & \cellcolor{mygray}Topographic (3) & \cellcolor{mygray}\xmark & \cellcolor{mygray}\xmark & \cellcolor{mygray}\xmark & \cellcolor{mygray}\xmark & \cellcolor{mygray}\xmark & \cellcolor{mygray}\cmark & \cellcolor{mygray}\cmark & \cellcolor{mygray}\cmark &  \cellcolor{mygray}\cmark  & \cellcolor{mygray}\cmark\\
        & Location (2) & \xmark & \cmark & \xmark & \xmark & \xmark & \xmark & \cmark & \cmark & \cmark  & \cmark\\
        & \cellcolor{mygray}Human footprint (9)  & \cellcolor{mygray}\xmark & \cellcolor{mygray}\xmark & \cellcolor{mygray}\xmark & \cellcolor{mygray} \xmark & \cellcolor{mygray}\xmark & \cellcolor{mygray}\xmark & \cellcolor{mygray}\xmark & \cellcolor{mygray}\cmark & \cellcolor{mygray}\cmark  & \cellcolor{mygray}\cmark \\
        & Plot metadata (20) & \xmark & \xmark & \xmark & \xmark & \xmark & \xmark & \xmark & \xmark & \cmark  & \cmark \\
        & \cellcolor{mygray}Satellite image features & \cellcolor{mygray}\xmark & \cellcolor{mygray}\xmark & \cellcolor{mygray}\cmark & \cellcolor{mygray}\xmark & \cellcolor{mygray}\xmark & \cellcolor{mygray}\xmark & \cellcolor{mygray}\xmark & \cellcolor{mygray}\xmark & \cellcolor{mygray}\xmark  & \cellcolor{mygray}\cmark \\
         \hline
        \multirow{9}{*}{\rotatebox[origin=c]{90}{\textbf{Method}}}
        & \textbf{Dummy}:        &  &  &  &  &  &  &  &  &  &  \\
        & All-Zero Predictor       & 0.135 & 0.023 & 0.035 & 0.029 & 0.025 & 0.026 & 0.025 & 0.028 & 0.025 & 0.022    \\
        & \textbf{Imputing}:        &  &  &  &  &  &  &  &  &  &  \\
        & Mean       & 0.133 & 0.026 & 0.031 & 0.027 & 0.023 & 0.022 & 0.016 & 0.018 & 0.016 &
       \textbf{0.000}    \\
        & Median       & 0.131 & 0.032 & 0.034 & 0.028 & 0.023 & 0.022 & 0.015 & 0.015 & 0.013 &
       0.003 \\
        & Marginal        & 0.135 & 0.023 & 0.035 & 0.028 & 0.024 & 0.024 & 0.021 & 0.023 & 0.017 & \textbf{0.000} \\ 
        & Conditional        & 0.129 & \textbf{0.009} & 0.034 & 0.008 & 0.007 & 0.007 & 0.006 & 0.008 & \textbf{0.005} & \textbf{0.000} \\\cline{2-12}
        & \textbf{Masking}:        &  &  &  &  &  &  &  &  &  &  \\
        & MaskSDM (ours)    & \textbf{0.028} & 0.036 & \textbf{0.013} & \textbf{0.006} & \textbf{0.006} & \textbf{0.005} & \textbf{0.005} & \textbf{0.005} & \textbf{0.005} & 0.004 \\ \hline
    \end{tabular}%
    \caption{Mean squared difference between test set predictions of the oracle and other baselines across species. A smaller difference indicates that the baseline's predictions are closer to those of the oracle. The dummy baseline consistently predicts zero. Bold values indicate the baseline with the smallest difference in each column.}
    \label{tab:diff_oracle}
\end{table}

In Table \ref{tab:diff_oracle}, we present the mean squared difference between test set predictions of the oracle and the baselines across species. We observe that MaskSDM achieves the smallest squared difference with the oracle compared to other baselines, except when using only location data or all predictors.
In the first case, it is unsurprising that the conditional imputation baseline performs best, as it can effectively approximate missing predictor values from neighboring observations. Additionally, all models showed significant improvements when trained longer on location data, suggesting that MaskSDM could further reduce this gap with extended training.
In the second case, the observed zero difference is expected, as the same model is used to generate both the oracle predictions and those of the mean, marginal, and conditional imputation baselines, with only a small number of missing values. Nevertheless, these results further highlight the ability of MaskSDM to effectively approximate a model trained with fewer predictors.

\subsection{Baseline prediction maps comparison}

\begin{figure}
    \centering
    \includegraphics[width=1\linewidth]{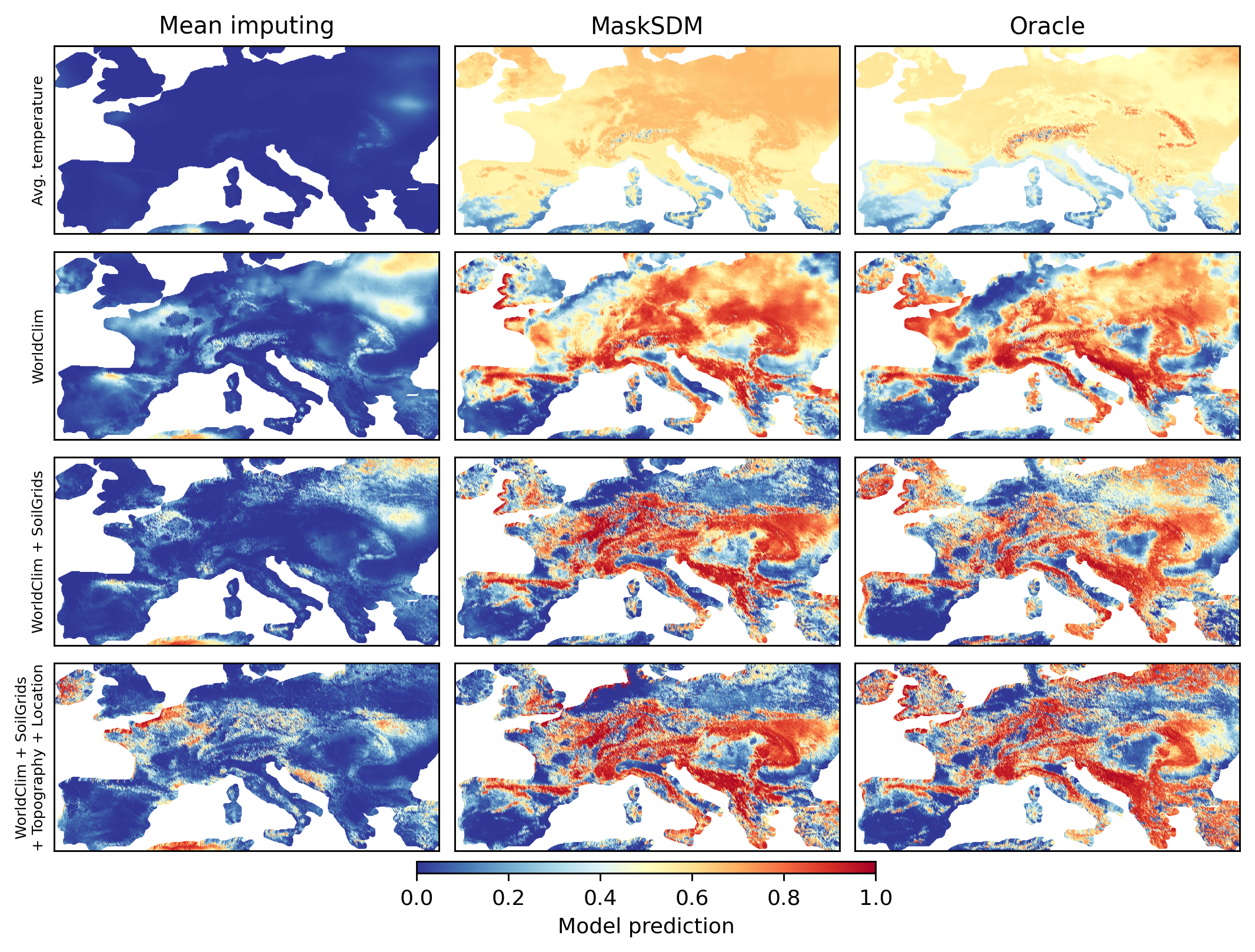}
    \caption{Comparison of predicted suitability maps for \textit{Anthyllis vulneraria} using different baselines and varying subsets of input variables.}
    \label{fig:oracle_maps_comparison}
\end{figure}

Fig. \ref{fig:oracle_maps_comparison} shows the suitability prediction maps for \textit{Anthyllis vulneraria} generated using the mean imputing baseline, MaskSDM, and the oracle. The mean imputation baseline produces poor predictions, struggling to account for missing variables effectively. In contrast, the predictions from MaskSDM closely resemble those from the oracle. The remaining minor differences may be attributed to the stochastic nature of model training.

\subsection{Performance of alternative modeling approaches}
\label{sec:appendix_architecture}

We investigate whether alternative modeling approaches commonly used in SDMs can match or outperform the FTTransformer-based MaskSDM, particularly in scenarios where all predictors are available. This analysis helps to assess the sensitivity of the results to changes in the modeling architecture. We compare MaskSDM to single-species, non–deep learning models, including a basic linear model and a linear model with Maxent-style feature augmentation \citep{maxent}, following the implementation of \cite{Anderson2023} and applying it to the same presence-absence data. Two sets of augmented features are tested: the first includes linear, hinge, and product terms, while the second additionally incorporates threshold and quadratic features.
We also evaluate two deep learning architectures for tabular data: a Multilayer Perceptron (MLP) and a Residual Network (ResNet), as described in \cite{gorishniy2021revisiting}, and we vary the number of blocks in each. For all these baselines, missing predictors are imputed using the mean value. Finally, to examine the sensitivity of MaskSDM to model depth, we include a variant with only three transformer blocks. The parameters of all these models are learned using the same training procedure as MaskSDM, except that masking is not applied for the imputation baselines.

\begin{table}
    \renewcommand{\arraystretch}{\stretchtable}
    \setlength{\arrayrulewidth}{0.2pt}
    \setlength\tabcolsep{5pt}
    \centering
    \small
    \begin{tabular}
        {|l|lcrc|cccccc|}
        \hline
        \multirow{7}{*}{\rotatebox[origin=c]{90}{\textbf{Predictors (\#)}}} 
        & Avg. Temperature (1) &  &  & & \cmark & \xmark & \cmark & \cmark & \cmark  & \cmark \\
        & \cellcolor{mygray}WorldClim (19) & \cellcolor{mygray} & \cellcolor{mygray} &\cellcolor{mygray} & \cellcolor{mygray}\xmark & \cellcolor{mygray}\xmark & \cellcolor{mygray}\cmark & \cellcolor{mygray}\cmark & \cellcolor{mygray}\cmark & \cellcolor{mygray}\cmark \\
        & SoilGrids (8) & & & & \xmark & \xmark & \xmark & \cmark & \cmark  & \cmark  \\
        & \cellcolor{mygray}Topographic (3) & \cellcolor{mygray} & \cellcolor{mygray} & \cellcolor{mygray} & \cellcolor{mygray}\xmark & \cellcolor{mygray}\xmark & \cellcolor{mygray}\xmark & \cellcolor{mygray}\xmark & \cellcolor{mygray}\cmark  &\cellcolor{mygray}\cmark  \\
        & Location (2) & & & & \xmark & \cmark & \xmark & \xmark & \cmark  &\cmark \\
        & \cellcolor{mygray}Human footprint (9) & \cellcolor{mygray} & \cellcolor{mygray} &  \cellcolor{mygray} & \cellcolor{mygray}\xmark & \cellcolor{mygray}\xmark & \cellcolor{mygray}\xmark & \cellcolor{mygray}\xmark & \cellcolor{mygray}\xmark  & \cellcolor{mygray}\cmark  \\
        & Plot metadata (20) &  & & & \xmark & \xmark & \xmark & \xmark & \xmark  & \cmark \\
         \hline
        \multirow{12}{*}{\rotatebox[origin=c]{90}{\textbf{Method}}} 
        &   &   variant  & \#parameters  & inf. time (s)  &  &  &  &  &  &  \\
        & \textbf{Mean imputing}:        &  &  &  &  &  &  & &  &  \\
        & Linear    &   & 789,756 & 0.31 & 69.8 & 78.2 & 88.4 & 93.4 & 95.8 & 96.2\\
        & Maxent & LHP       & 38,086,620 & 0.53 & 81.4 & 87.7 & 94.6 & 95.8 & 96.6 & 95.7 \\
        & Maxent & LHPTQ   & 46,633,818 & 0.56 & 83.6 & 90.3 & 95.1 & 95.9 & 96.6 & 96.0  \\ 
        & MLP    & 7 blocks  & 2,692,674 & 0.28 & 68.7 & 74.2 & 86.8 & 87.8 & 97.0 & 97.4  \\
        & MLP    & 21 blocks  & 3,211,458 & 0.29 & 41.0 & 63.3 & 78.3 & 78.6 & 91.5 & 92.3  \\
        & ResNet  & 7 blocks     & 2,992,194 & 0.29 & 67.6 & 78.6 & 90.9 & 93.8 & 97.3 & 97.7 \\
        & ResNet    & 21 blocks  & 4,035,138 & 0.31 & 70.6 & 80.1 & 91.9 & 94.4 & 97.4 & 97.8  \\
        \cline{2-11}
        & \textbf{Masking}:        &  &  &  &  &  &  & &  & \\
        & MaskSDM  & 3 blocks & 4,390,578 & 0.70 & \textbf{85.5} & 95.9 & 97.0 & 97.3 & \textbf{97.6} & 97.9 \\
        & MaskSDM  & 7 blocks & 5,381,554 & 1.10 & 85.1 & \textbf{96.0} & \textbf{97.1} & \textbf{97.4} & \textbf{97.6} & \textbf{98.0} \\
        \hline
    \end{tabular}%
    \caption{AUC performance of different deep learning architectures and modeling approaches on the \textbf{sPlotOpen dataset}. The inference time (in seconds) is the total time required to generate predictions for the entire test set. The setup of the methods is detailed in the text. The best-performing method in each column is highlighted in bold.}
    \label{tab:architecture}
\end{table}

Table \ref{tab:architecture} reports the AUC performance of these models, along with their number of trainable parameters and inference time for the entire test set. MaskSDM consistently outperforms the other approaches, with ResNet achieving the closest performance, particularly when nearly all predictors are available. We also observe that the more lightweight version of MaskSDM (3 versus 7 blocks) does not significantly degrade performance. In future work, we plan to further optimize MaskSDM’s hyperparameters to identify an optimal trade-off between computational efficiency and predictive accuracy.

\subsection{Performance on the GeoPlant dataset}
\label{sec:appendix_geoplant}

\begin{table}
    \renewcommand{\arraystretch}{\stretchtable}
    \setlength{\arrayrulewidth}{0.2pt}
    \setlength\tabcolsep{5pt}
    \centering
    \small
    \begin{tabular}
        {|l|lcrc|cccccc|}
        \hline
        \multirow{7}{*}{\rotatebox[origin=c]{90}{\textbf{Predictors (\#)}}} 
        & Avg. Temperature (1) &  &  & & \cmark & \xmark & \cmark & \cmark & \cmark  & \cmark \\
        & \cellcolor{mygray}WorldClim (19) & \cellcolor{mygray} & \cellcolor{mygray} &\cellcolor{mygray} & \cellcolor{mygray}\xmark & \cellcolor{mygray}\xmark & \cellcolor{mygray}\cmark & \cellcolor{mygray}\cmark & \cellcolor{mygray}\cmark & \cellcolor{mygray}\cmark \\
        & SoilGrids (8) & & & & \xmark & \xmark & \xmark & \cmark & \cmark  & \cmark  \\
        & \cellcolor{mygray}Elevation (1) & \cellcolor{mygray} & \cellcolor{mygray} & \cellcolor{mygray} & \cellcolor{mygray}\xmark & \cellcolor{mygray}\xmark & \cellcolor{mygray}\xmark & \cellcolor{mygray}\xmark & \cellcolor{mygray}\cmark  &\cellcolor{mygray}\cmark  \\
        & Location (2) & & & & \xmark & \cmark & \xmark & \xmark & \cmark  &\cmark \\
        & \cellcolor{mygray}Human footprint (9) & \cellcolor{mygray} & \cellcolor{mygray} &  \cellcolor{mygray} & \cellcolor{mygray}\xmark & \cellcolor{mygray}\xmark & \cellcolor{mygray}\xmark & \cellcolor{mygray}\xmark & \cellcolor{mygray}\xmark  & \cellcolor{mygray}\cmark  \\
        & Plot metadata (2) &  & & & \xmark & \xmark & \xmark & \xmark & \xmark  & \cmark \\
         \hline
        \multirow{12}{*}{\rotatebox[origin=c]{90}{\textbf{Method}}} 
        &   &   variant  & \#parameters  & inf. time (s)  &  &  &  &  &  &  \\
        & \textbf{Mean imputing}:        &  &  &  &  &  &  & &  &  \\
        & Linear    &   & 95,675 & 0.26 & 65.6 & 64.9 & 82.6 & 85.0 & 86.3 & 87.4 \\
        & Maxent & LHP       & 3,693,500 & 0.37 & 71.5 & 67.8 & 82.9 & 83.6 & 85.5 & 85.4 \\
        & Maxent & LHPTQ   & 4,721,450 & 0.43 & 72.1 & 70.2 & 83.1 & 83.5 & 85.8 & 85.6  \\ 
        & MLP    & 7 blocks  & 660,017 & 0.26 & 65.0 & 64.8 & 83.3 & 85.7 & 87.9 & 88.4 \\
        & MLP    & 21 blocks  & 1,178,801 & 0.24 & 57.5 & 53.8 & 72.9 & 78.1 & 80.8 & 82.5 \\
        & ResNet  & 7 blocks     & 959,537 & 0.38 & 66.9 & 67.9 & 84.2 & 86.4 & 88.4 & 89.3 \\
        & ResNet    & 21 blocks  & 2,002,481 & 0.27 & 68.2 & 63.3 & 85.5 & 87.1 & 88.6 & 89.3 \\
        \cline{2-11}
        & \textbf{Masking}:        &  &  &  &  &  &  & &  & \\
        & MaskSDM  & 3 blocks & 2,006,801 & 0.45 & 79.6 & \textbf{80.9} & 87.4 & 88.4 & 88.7 & \textbf{89.7} \\
        & MaskSDM  & 7 blocks & 2,997,777 & 0.75 & \textbf{80.0} & 80.8 & \textbf{87.7} & \textbf{88.6} & \textbf{88.8} & 89.5 \\
        \hline
    \end{tabular}%
    \caption{AUC performance of different deep learning architectures and modeling approaches on the \textbf{GeoPlant dataset}. The inference time (in seconds) is the total time required to generate predictions for the entire test set. AThe setup of the methods is detailed in the text. The best-performing method in each column is highlighted in bold.}
    \label{tab:geoplant}
\end{table}

In this section, we demonstrate that MaskSDM generalizes well to other datasets. To this end, we evaluate MaskSDM performance on the GeoPlant dataset \citep{picek2024geoplant}, which, like sPlotOpen, consists of vegetation plots, but is restricted to Europe. As before, we focus on presence-absence data and use the 88,987 validation plots provided by the authors. We apply the same filtering procedure used with sPlotOpen, retaining only species with at least 20 occurrences, resulting in 2225 modeled species. The data is split using the same block cross-validation strategy  \citep{roberts2017cross}. We use the tabular predictors available in GeoPlant, selecting those that align with the setup of sPlotOpen. Specifically, we include variables from WorldClim, SoilGrids, elevation, geographic coordinates, human footprint, and metadata (plot size and location uncertainty). All predictors are directly available for each plot in the dataset. We evaluate the same set of alternative modeling approaches described in Appendix \ref{sec:appendix_architecture}. The results in Table \ref{tab:geoplant} show trends very similar to those observed with sPlotOpen: MaskSDM consistently outperforms the other models and remains robust across architectural variations. These findings highlight the potential of MaskSDM to be implemented across multiple datasets. 

\subsection{Additional prediction maps}
\label{sec:appendix_prediction_maps}

\begin{figure}
    \centering
    \includegraphics[width=0.765\linewidth]{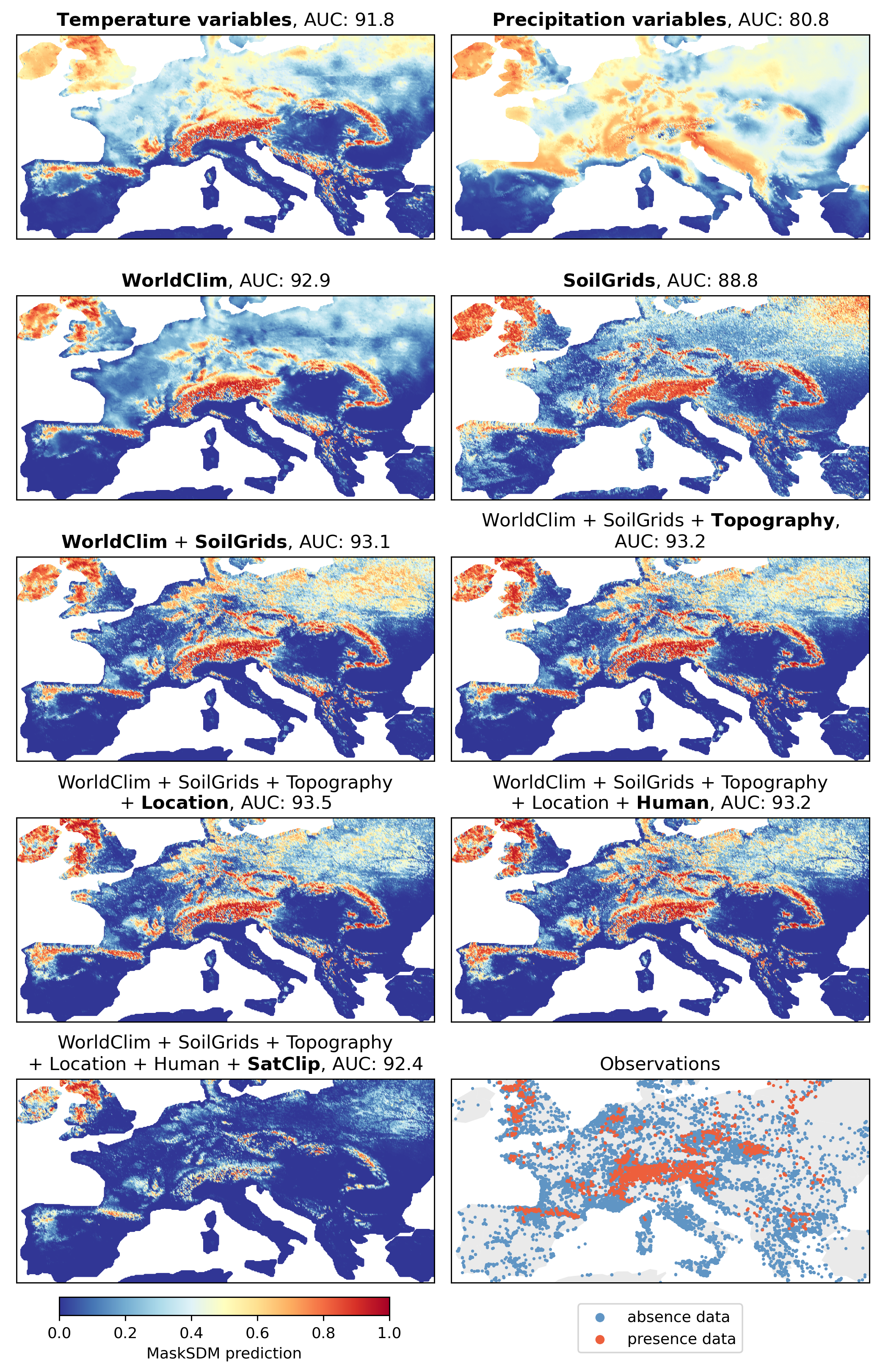}
    \caption{MaskSDM predicted suitability maps for the European blueberry (\textit{Vaccinium myrtillus}) using different subsets of input variables. For each subset, we report the corresponding AUC obtained for \textit{V. myrtillus} in the test set. The bottom-right panel shows the geographic distribution of observations, with presence data marked in red and absence data in blue.}
    \label{fig:vaccinium_myrtillus}
\end{figure}

\begin{figure}
    \centering
    \includegraphics[width=0.765\linewidth]{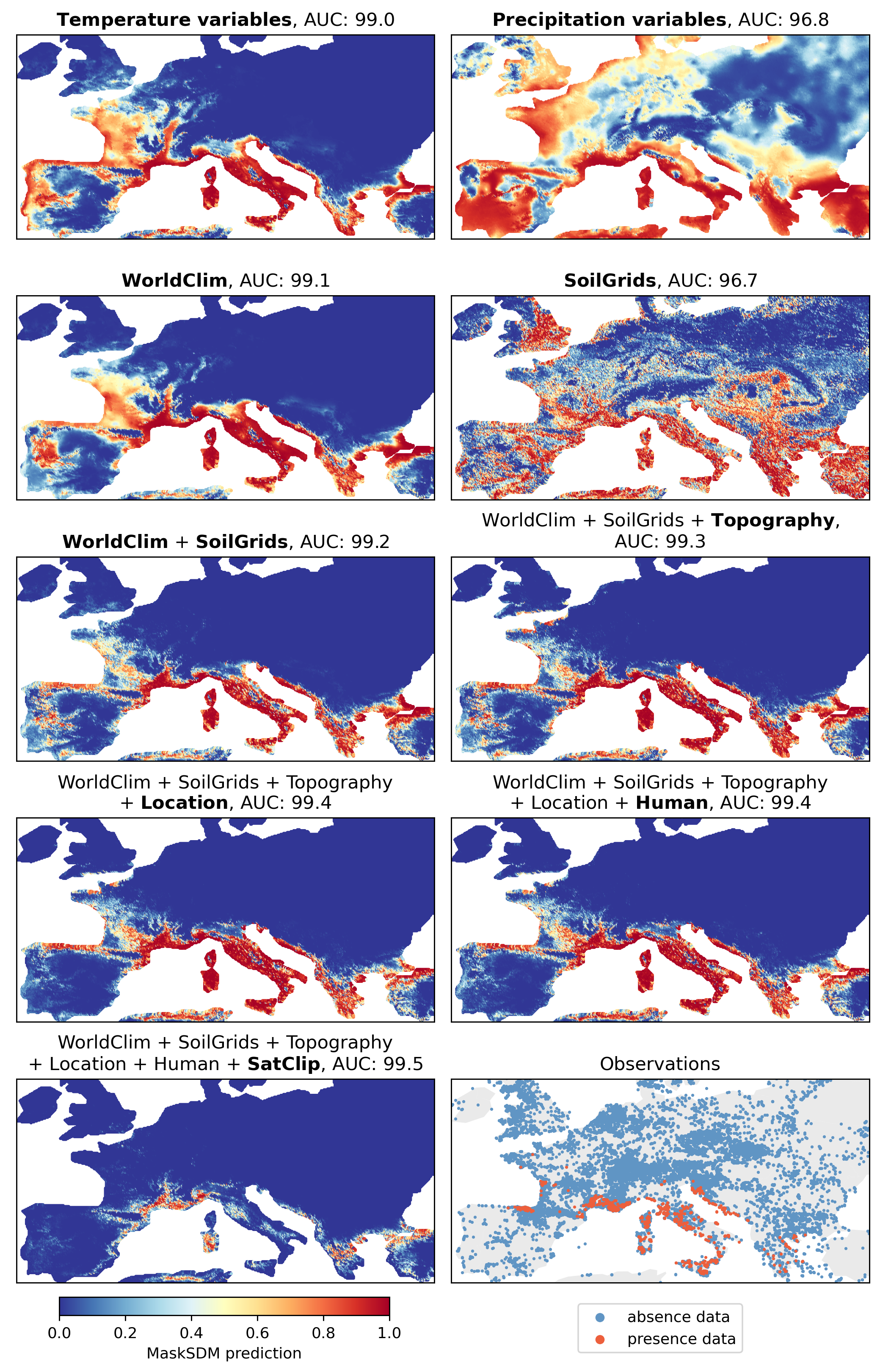}
    \caption{MaskSDM predicted suitability maps for the holm oak (\textit{Quercus ilex}) using different subsets of input variables. For each subset, we report the corresponding AUC obtained for \textit{Q. ilex} in the test set. The bottom-right panel shows the geographic distribution of observations, with presence data marked in red and absence data in blue.}
    \label{fig:quercus_ilex}
\end{figure}

Figs \ref{fig:vaccinium_myrtillus} and \ref{fig:quercus_ilex} present prediction maps for \textit{Vaccinium myrtillus} and \textit{Quercus ilex} respectively.

\end{document}